\documentclass[12pt]{article}

\usepackage{setspace}
\usepackage{arxiv}
\usepackage[utf8]{inputenc} 
\usepackage[T1]{fontenc}    
\usepackage{hyperref}       
\usepackage{url}            
\usepackage{booktabs}       
\usepackage{amsfonts}       
\usepackage{nicefrac}       
\usepackage{microtype}      
\usepackage{lipsum}		
\usepackage{graphicx}
\usepackage[numbers,sort&compress]{natbib}
\usepackage{doi}
\usepackage{amsmath} 
\usepackage{booktabs,makecell,multirow}
\usepackage{float}

\title{FuXiWeather2: Learning accurate atmospheric state estimation for operational global weather forecasting}

\author{
	Xiaoze Xu$^{1\dagger}$,
	Xiuyu Sun$^{1\dagger}$,
    Songling Zhu$^{1}$,
    Xiaohui Zhong$^{2,1,3}$, 
    Yuanqing Huang$^{1}$,
    Zijian Zhu$^{2,1}$, \\
    \textbf{Jun Liu}$^{2,1}$,
	\textbf{Hao Li}$^{2,1,3\ast}$ 
    \and
	\small$^{1}$Shanghai Academy of Artificial Intelligence for Science, Shanghai, \& 200232, China.\and
	\small$^{2}$Artificial Intelligence Innovation and Incubation Institute, Fudan University, Shanghai \& 200433, China.\and
    \small$^{3}$FuXi Intelligent Computing Technology Co., Ltd., Shanghai \& 200233, China.\and
	\small$^\ast$Corresponding author. Email: lihao\_lh@fudan.edu.cn \and 
	\small$^\dagger$These authors contributed equally to this work.
}

\renewcommand{\shorttitle}{\textit{FuXiWeather2}}

\begin{document}
\maketitle

\begin{abstract}
Numerical weather prediction has long been constrained by the computational bottlenecks inherent in data assimilation and numerical modeling. While machine learning has accelerated forecasting, existing models largely serve as "emulators of reanalysis products," thereby retaining their systematic biases and operational latencies. Here, we present FuXiWeather2, a unified end-to-end neural framework for assimilation and forecasting. We align training objectives directly with a combination of real-world observations and reanalysis data, enabling the framework to effectively rectify inherent errors within reanalysis products. To address the distribution shift between NWP-derived background inputs during training and self-generated backgrounds during deployment, we introduce a recursive unrolling training method to enhance the precision and stability of analysis generation. Furthermore, our model is trained on a hybrid dataset of raw and simulated observations to mitigate the impact of observational distribution inconsistency. FuXiWeather2 generates high-resolution ($0.25^{\circ}$) global analysis fields and 10-day forecasts within minutes. The analysis fields surpass the NCEP-GFS across most variables and demonstrate superior accuracy over both ERA5 and the ECMWF-HRES system in lower-tropospheric and surface variables. These high-quality analysis fields drive deterministic forecasts that exceed the skill of the HRES system in 91\% of evaluated metrics. Additionally, its outstanding performance in typhoon track prediction underscores its practical value for rapid response to extreme weather events. The FuXiWeather2 analysis dataset is available at \url{https://doi.org/10.5281/zenodo.18872728}.

\end{abstract}

\keywords{ Weather Prediction \and Machine Learning \and Data Assimilation \and FuXi \and FuXiWeather2 }

\newpage
\section{Introduction}

For decades, numerical weather prediction (NWP) systems have served as the essential infrastructure for global disaster resilience and socioeconomic planning \cite{dutton2002opportunities,grasso2011early,lazo2011us,rogers2011costs,foley2012current}. The operational pipeline begins with data assimilation (DA), optimally blending tens of millions of heterogeneous observations with short-range forecasts (background) to estimate the current atmospheric state. This estimated state (analysis) serves as the initial condition for the prediction model, which drives atmospheric evolution by solving equations for large-scale fluid motion and thermodynamic processes, complemented by physical parameterization schemes \cite{pu2019numerical}.

However, escalating demands for higher-resolution and near-real-time forecasts are pushing traditional NWP systems to their operational limits \cite{bauer2015quiet,Michalakes2020,bauer2024}. 
Recently, state-of-the-art neural weather prediction models have demonstrated the capacity to match or exceed the accuracy of numerical models with thousand-fold speedups \cite{Chen2023,graphcast,pangu2023,Kochkov2024,lang2024aifs,Seeds2024,Chen2025,Price2025,fuxiens2025}. 
Despite these advances, such models are essentially reanalysis emulators rather than standalone forecasting systems. They are designed to map NWP analysis fields, typically derived from the ECMWF Reanalysis v5 (ERA5) dataset \cite{hersbach2020era5}, to future states, thereby bypassing the fundamental challenge of assimilating raw, irregular observations. This creates two critical limitations: (1) 
training exclusively on reanalysis products inherits NWP's systematic deficiencies, such as near-surface and precipitation errors \cite{sandu2020addressing,jiang2021evaluation,lavers2022evaluation,wang2024antarctic};
and (2) relying on operational analysis for initialization keeps them tethered to the latency of traditional DA pipelines \cite{haseler2004ear,bauer2015quiet}. 
To truly revolutionize global forecasting, a transition from modular emulation to an end-to-end neural framework is imperative.

Establishing such neural weather prediction systems faces fundamental hurdles, primarily in generating accurate analysis fields. Existing approaches generally follow two paradigms. The first maps observations directly to atmospheric states, discarding background information to simplify training \cite{alexe2024graphdop,zhao2024omg,allen2025,gupta2026healda}. However, lacking the stabilizing dynamics and thermodynamics of previous forecasts, these snapshot-based methods struggle to maintain precision in observation-sparse regions \cite{sun2025}, yielding coarse, low-fidelity analysis fields \cite{allen2025,gupta2026healda}. 
The second paradigm adheres to the traditional cyclic DA framework by integrating background priors with observations \cite{huang2024diffda,xu2025,sun2025,xiang2025adaf}. 
Despite improved stability, two substantial inconsistencies limit performance:
One is the inconsistency in background field generation (exposure bias). These methods are typically trained using reanalysis-driven background fields but must operate on self-generated background fields during deployment \cite{huang2024diffda,sun2025}. 
The other is the inconsistency in observational distribution. Due to the continuous evolution of observing systems \cite{bauer2015quiet,eyre2020part1,eyre2022part2}, a significant discrepancy exists between the observations available during model training and those used in operational deployment \cite{alexe2024graphdop}, particularly concerning newly deployed satellite sensors.
Recent benchmarks indicate that the analysis fields produced by state-of-the-art neural weather prediction systems exhibit errors comparable to those of 36-hour operational NWP forecasts \cite{sun2025,gupta2026healda}, failing to meet stringent operational standards. 
Most critically, these models remain constrained by using reanalysis data as their sole training targets. Consequently, achieving a competitive, fully autonomous neural weather prediction system has remained an elusive goal until now.

Here, we present FuXiWeather2, an end-to-end neural weather prediction framework bridging the gap between raw observations and global weather forecasting. By integrating multi-source remote sensing and in-situ data, it generates high-resolution ($0.25^{\circ}$) global analysis fields and 10-day forecasts within minutes. 
Our approach introduces three core innovations to address the fundamental challenges of AI-based analysis generation:
First, we overcome the limitations of pure reanalysis emulation. By anchoring the training objective to combined raw observations and reanalysis, FuXiWeather2 maintains thermodynamic and dynamical constraints while rectifying errors against discrete ground-truth observations.
Second, we address the inconsistency in background field generation through a recursive unrolled training paradigm. Unlike disjoint approaches \cite{huang2024diffda,allen2025,sun2025,xu2025,gupta2026healda}, our fully differentiable framework unifies assimilation and forecasting. Ingesting self-generated forecasts during training exposes the model to accumulating errors, forcing robust error correction and ensuring long-term stability.
Third, to mitigate the observational distribution inconsistency, we utilize physical forward models to generate simulated historical observations. 
The integration of raw and simulated data alleviates the non-stationarity of the observing system, providing the model with a stable and consistent training dataset spanning nearly five years.

In summary, FuXiWeather2 establishes a new frontier for global weather modeling. Its analysis fields surpass those of the National Centers for Environmental Prediction (NCEP) Global Forecast System (GFS) across nearly all variables and outperform both ERA5 and the European Centre for Medium-Range Weather Forecasts (ECMWF) High-Resolution (HRES) system in lower-troposphere and surface variables. Driven by these high-quality analysis fields, our deterministic forecasts outperform the HRES baseline in 91\% of the evaluated metrics, with the effective lead time for Z500 reaching 9.75 days. Notably, the system achieves superior typhoon track prediction accuracy compared to HRES while drastically reducing the inference latency to minutes, providing critical support for timely early warnings of extreme events. Finally, we have released an open-source, AI-based analysis dataset covering the period from July 2023 to June 2024 (\url{https://doi.org/10.5281/zenodo.18872728}).
\section{FuXiWeather2}\label{sec:fuxiweather2}
\begin{figure}[ht]
  \centering
  \includegraphics[width=1.0\textwidth]{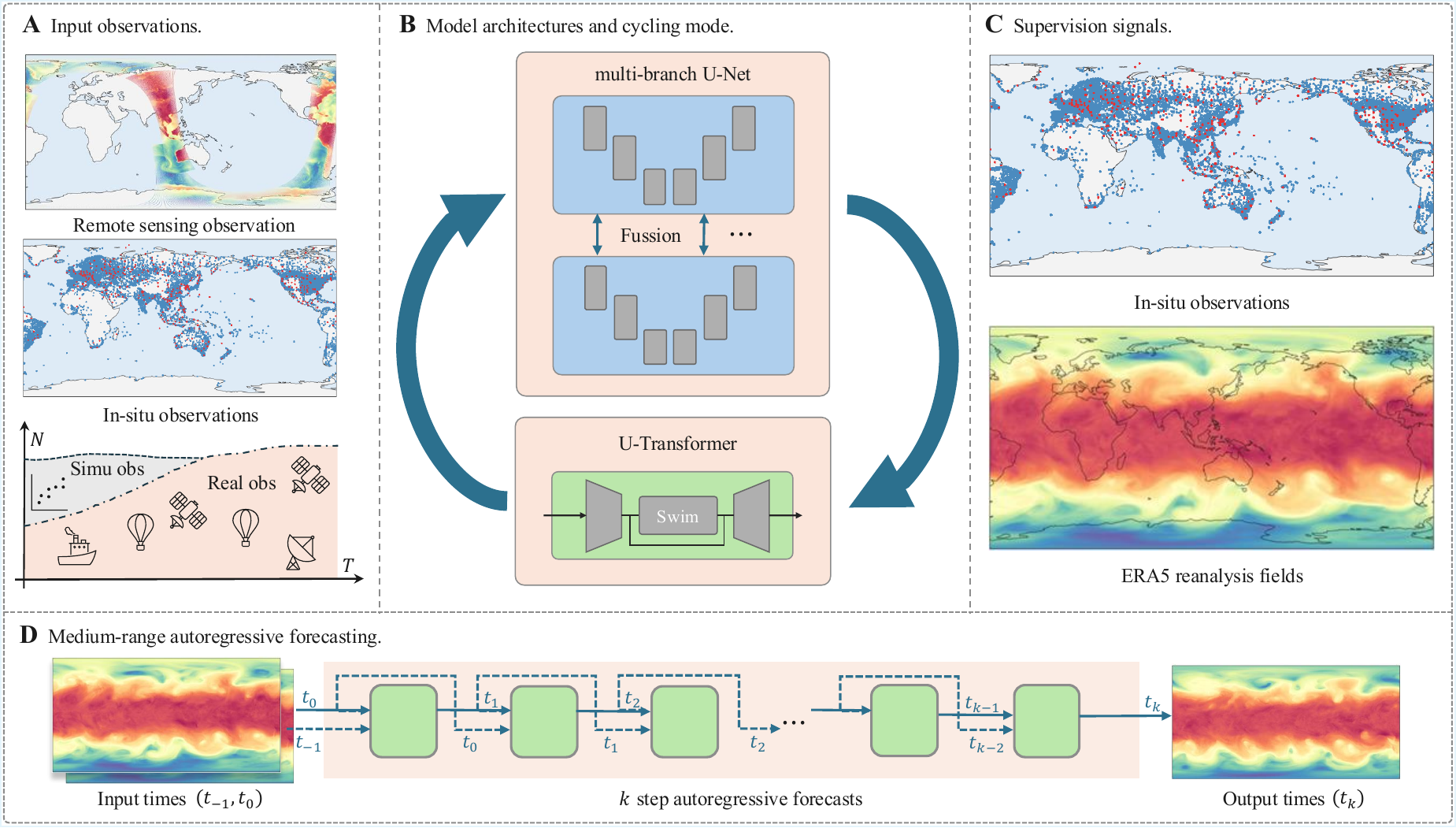}
  \caption{ \textbf{Overview of the FuXiWeather2 system}. (A) Input observations. The system ingests multi-source data, including remote sensing and in-situ measurements. A hybrid dataset, combining real-world observations and physical simulations, is utilized during training, while only real-world observations are used during inference. (B) Model architectures and cycling mode. The system consists of a multi-branch U-Net for data assimilation (DA) and a Swin U-Transformer for forec
  asting. The DA module integrates observations with short-range background forecast to generate analysis, which then drive the forecast module to produce the next-step background. Both modules operate in an interleaved cycling manner, continuously generating stable analysis fields. The entire process is optimized via a recursive unrolled end-to-end training paradigm. (C) Supervision signals. During training, FuXiWeather2 employs dual supervision, leveraging both global reanalysis products and discrete in-situ observations. (D) Medium-range autoregressive forecasting. Initialized by the analysis, the forecast module generates 10-day global forecasts through an autoregressive approach.}
  \label{fig:framework}
\end{figure}

FuXiWeather2 is an end-to-end neural weather prediction framework designed for the seamless integration of DA and forecasting. Structurally, the system comprises two specialized deep neural networks: a multi-branch convolutional U-Net for DA \cite{sun2025} and an autoregressive Swin-Transformer for forecasting \cite{Chen2023}. 
By directly ingesting Level-1 remote sensing data and in-situ measurements, the system operates independently of external NWP products during deployment, functioning as a fully autonomous forecasting system. Fig. \ref{fig:framework} provides an overview of the proposed framework, and a comprehensive list of atmospheric and observational variables is detailed in Table \ref{tab:sup_module_variables}.
\begin{table}[ht]
\centering
\caption{\textbf{Data utilized in the assimilation and forecast modules of FuXiWeather2.} This table summarizes the data categories, their corresponding functional modules, and their roles as either inputs (In) or outputs (Out).}
\small
\label{tab:sup_module_variables}
\begin{tabular}{llcc}
\toprule
\textbf{Category} & \textbf{Variable Name (Abbr.)} & \textbf{Module} & \textbf{I/O Role} \\
\midrule
\multirow{5}{*}{\textbf{Upper-air\textsuperscript{*}}} 
 & Geopotential ($Z$) & \multirow{5}{*}{Assim \& Fcst} & \multirow{5}{*}{In \& Out} \\
 & Temperature ($T$) & & \\
 & U-wind component ($U$) & & \\
 & V-wind component ($V$) & & \\
 & Relative humidity ($R$) & & \\
\midrule
\multirow{5}{*}{\textbf{Surface}} 
 & 2-meter temperature ($T_{2m}$) & \multirow{5}{*}{Assim \& Fcst} & \multirow{5}{*}{In \& Out} \\
 & Mean sea-level pressure ($MSLP$) & & \\
 & 10-meter U-wind component ($U_{10m}$) & & \\
 & 10-meter V-wind component ($V_{10m}$) & & \\
 & Total precipitation ($TP$) & & \\
\midrule
\multirow{9}{*}{\textbf{Satellite}} 
 & \makecell[l]{Advanced Technology Microwave Sounder (ATMS)} & \multirow{9}{*}{Assim} & \multirow{9}{*}{In}\\
 & \makecell[l]{Microwave Humidity Sounder-II (MWHS-II)} & & \\
 & \makecell[l]{Microwave Temperature Sounder-III (MWTS-III)} & & \\
 & \makecell[l]{Advanced Microwave Sounding Unit-A (AMSU-A)} & & \\
 & \makecell[l]{Microwave Humidity Sounder (MHS)} & & \\
 & \makecell[l]{Special Sensor Microwave-Imager/Sounder (SSMIS)} & & \\
 & \makecell[l]{High-resolution Infra Red Sounder (HIRS)} & & \\
 & \makecell[l]{Infrared Atmospheric Sounding Interferometer (IASI)} & & \\
 & \makecell[l]{GNSS Radio Occultation (GNSS-RO)} & & \\
\midrule
\textbf{In-situ} & Land station, Radiosonde, Marine platform & Assim & In \\
\midrule
\multirow{3}{*}{\textbf{Geographical}} 
 & Land-sea mask ($LSM$), Lake cover ($LC$) & \multirow{3}{*}{Assim} & \multirow{3}{*}{In} \\
 & Surface geopotential ($SG$), Vegetation cover\textsuperscript{**} ($VC$) & & \\ 
 & Latitude ($LAT$), Longitude ($LON$) & & \\
\midrule
\multirow{1}{*}{\textbf{Temporal}} 
 & Hour of day ($HOUR$), Day of year ($DOY$) & \multirow{1}{*}{Fcst} & \multirow{1}{*}{In} \\
\bottomrule
\end{tabular}
\begin{flushleft}
\footnotesize \textsuperscript{*} Upper-air atmospheric variables are distributed across 13 vertical pressure levels: 50, 100, 150, 200, 250, 300, 400, 500, 600, 700, 850, 925, and 1000 hPa. \\
\footnotesize \textsuperscript{**} Vegetation cover includes Low vegetation cover (LVC), High vegetation cover (HVC), Type of low vegetation (TLV), and Type of high vegetation (THV).
\end{flushleft}
\end{table}
The operational workflow of FuXiWeather2 is structured into two core stages: First, all observational data are mapped onto a regular 0.25$^\circ$ grid, matching the resolution of the background fields. To accommodate the distinct spatiotemporal characteristics of heterogeneous data, we categorize observations into two groups and implement specialized encoding schemes for each (see section \ref{sec_obs_encode}). Second, FuXiWeather2 employs a cycling paradigm to generate analysis fields, where the assimilation module estimates the current atmospheric state (the analysis) by integrating encoded observations with a 6-hour forecast (the background) generated by the forecasting module. The generated analysis then serves as the initial condition for the forecasting module to generate the background for the next cycle step. For medium-range weather prediction, FuXiWeather2 follows a standard autoregressive approach, achieving long-term predictions by recursively feeding its own outputs back as inputs \cite{graphcast,Chen2023}.

Our ultimate objective is to achieve full operational autonomy, where the model sustains its own cycling process without reliance on external NWP analysis. This requires the model to master the complete workflow: initiating from an uninformative atmospheric state (e.g., zero-valued field) and incrementally building toward a stable cycling state through the assimilation of observations.
We propose a recursive unrolled training framework to simulate this process. However, training from zero-valued fields requires extensive unrolling steps (more than ten steps) to reach a steady state, which severely hampers convergence and inflates computational costs. To bypass this, we introduce a warm-start strategy that samples initial states from pre-generated forecasts of varying lead times. This exposes the model to diverse accumulated errors while significantly reducing the unrolling depth to a practical range of 4-6 steps (see section \ref{sec_unrolled_training}).
Furthermore, to mitigate the memory bottleneck caused by integrating vast observational data within the unrolled sequences, we design a pipeline-parallelism strategy. By distributing the constituent network modules across separate GPU devices, we enable the high-resolution (0.25$^\circ$) training of a unified, fully differentiable paradigm that was previously computationally infeasible (see section \ref{sec_compu_imple}).

To overcome the performance bottleneck imposed by training solely on reanalysis targets, we employ a joint latitude‑weighted L1 loss function that incorporates both ERA5 and in‑situ observations as learning targets  (see \ref{sec_train_obj}). This loss function is applied to the outputs of every forecasting and assimilation step within the recursive unrolled training.
FuXiWeather2 is trained on a hybrid dataset consisting of real-world observations and simulations from physical forward models spanning June 1, 2018, to June 1, 2023 (see section \ref{sec_forward_sim}). During the inference stage, the system operates entirely on real-world observations. 
Starting from a cold-start initialization on June 20, 2023, the system undergoes a continuous cycle of assimilation and forecasting. Following a 10-day spin-up period to reach a stable state, the formal evaluation is conducted over a full year from July 1, 2023, to June 30, 2024.
\subsection{Data Sources and Processing}
\subsubsection{Data Sources}
Building on their well-documented contributions to operational numerical weather prediction (NWP) systems \cite{eyre2022part2,samrat2025observation}, a representative suite of observational data sources was prioritized, primarily categorized into: (1) remote sensing observations, including microwave/infrared sounders and Global Navigation Satellite System (GNSS) radio occultation (RO) data; and (2) in-situ measurements from radiosondes, land stations, and marine platforms. To ensure that FuXiWeather2 operates as an autonomous system independent of external NWP priors, we strictly utilized Level-1 brightness temperatures from microwave/infrared sounders and refractivity data from GNSS-RO. Detailed configurations for all observation types are provided in Table \ref{tab:sub_obs_variables}. 
\begin{table}[ht]
\centering
\caption{ \textbf{Detailed specifications of the observational datasets assimilated in FuXiWeather2.} The system directly assimilates satellite Level-1 brightness temperatures and refractivity data, thereby avoiding reliance on retrieved products that may be influenced by external numerical weather prediction (NWP) priors. In-situ observations provide direct measurements. $t_0$ denotes the analysis time.}
\label{tab:sub_obs_variables}
\small
\begin{tabular}{llllll}
\toprule
\textbf{Category} & \textbf{Instrument} & \textbf{Platform} & \textbf{Provider} & \textbf{Assimilation Variable} & \textbf{Window (h)} \\
\midrule
\multirow{9}{*}{\textbf{Satellite}} 
 & ATMS & NOAA-20, NPP & NOAA & Brightness Temperature & $[t_0-3, t_0+5)$ \\
 & AMSU-A & \makecell[l]{NOAA-15/18/19,\\ METOP-B/C} & NOAA & Brightness Temperature & $[t_0-3, t_0+5)$ \\
 & MHS & \makecell[l]{NOAA-19,\\ METOP-B/C} & NOAA & Brightness Temperature & $[t_0-3, t_0+5)$ \\
 & MWTS-III\textsuperscript{*} & FY-3E & CMA & Brightness Temperature & $[t_0-3, t_0+5)$ \\
 & MWHS-II\textsuperscript{*} & FY-3E & CMA & Brightness Temperature & $[t_0-3, t_0+5)$ \\
 & SSMIS & DMSP-F17/F18 & NOAA & Brightness Temperature & $[t_0-3, t_0+5)$ \\
 & IASI & METOP-C & NOAA & Brightness Temperature & $[t_0-3, t_0+5)$ \\
 & HIRS-4 & NOAA-19 & NOAA & Brightness Temperature & $[t_0-3, t_0+5)$ \\
 & GNSS-RO\textsuperscript{*} & \makecell[l]{COSMIC-2,\\ METOP, FY-3, etc.} & \makecell[l]{CMA/\\NOAA} & Refractive & $[t_0-3, t_0+5)$ \\
\midrule
\multirow{3}{*}{\textbf{In situ}} 
 & Radiosonde & Global Network & NOAA & $T, U, V, Q, Z$ & $[t_0-3, t_0+3)$ \\
 & Land Station & Global Synop & NOAA & \makecell[l]{ $T_{2m}$, $R_{2m}$, $U_{10m}$, \\ $V_{10m}$, $SP$, $MSLP$} & $[t_0-3, t_0+5)$ \\
 & Marine & Ship / Buoy & NOAA & 
 \makecell[l]{$T_{2m}$, $R_{2m}$, $U_{10m}$,\\ $V_{10m}$, $MSLP$, $SST$ } & $[t_0-3, t_0+5)$ \\
\bottomrule
\end{tabular}
\begin{flushleft}
\footnotesize \textsuperscript{*} For GNSS-RO, MWHS-II, and MWTS-III, data prior to June 2023 are simulated from physical models, whereas data from June 2023 onward are real observations
\end{flushleft}
\end{table}
Global atmospheric state constraints are primarily provided by a diverse network of satellite observations, which ensure continuous coverage across regions where in-situ instruments are limited. We assimilated data from a comprehensive suite of instruments, including microwave sounders/imagers (AMSU-A, MHS, ATMS, MWHS-II, MWTS-III and SSMIS), and infrared sounders (HIRS-4, IASI). 
To address systematic biases arising from scanning-angle variations and cross-platform calibration discrepancies, we append the satellite zenith angle and platform identity to the input alongside the raw brightness temperatures. 
This design enables the neural network to learn implicit bias corrections from historical data, effectively obviating the platform-specific bias correction procedures typically required in traditional NWP. Taking ATMS as an illustrative example, the standardized input channel dimension is 25: it comprises 22 raw brightness temperature channels, concatenated with one satellite zenith angle channel and two one-hot encoding channels representing the satellite platform identity (e.g., distinguishing between NOAA-20 and NPP). For instruments deployed on multiple platforms (e.g., AMSU-A on five platforms, Table \ref{tab:sub_obs_variables}), the platform encoding expands to five one-hot encoding channels.

Radiosonde observations serve as anchor measurements, providing profiles of $T$, $R$, $U$, $V$, and $Z$ (geopotential is calculated by multiplying geopotential height by the standard gravity, $g=9.80665 m/s^2$). Data from 13 pressure levels consistent with the forecast models were adopted (details of these pressure levels are provided in Table \ref{tab:sup_module_variables}). However, these observations are predominantly land-based and confined to specific synoptic hours (00:00 and 12:00 UTC). To bridge these spatio-temporal gaps, we incorporate GNSS-RO data as a critical supplement, leveraging its near-unbiased characteristics in the middle and upper atmosphere \cite{wickert2005gps}. GNSS-RO offers globally distributed, high-vertical-resolution refractivity profiles that are essential for constraining the thermodynamic structure. In this study, refractivity profiles extending up to 50 km are uniformly sampled into 512 height layers \cite{sun2025}.

Near-surface in-situ observations are essential for constraining the state of the lower atmosphere. While satellite sounders provide vertical atmospheric profiles, their sensitivity is often limited in the lower troposphere. This study utilizes global land station observations, encompassing 2-meter temperature ($T_{2m}$), 2-meter relative humidity ($R_{2m}$), 10-meter U-wind component and 10-meter V-wind component ($U_{10m}, V_{10m}$), surface pressure ($P_s$), and mean sea level pressure ($MSLP$), provided alongside station elevation as auxiliary metadata.
Furthermore, to ensure adequate coverage over marine regions, we incorporate $T_{2m}$, $R_{2m}$, $U_{10m}$, $V_{10m}$, $MSLP$, and sea surface temperature ($SST$) data from the International Comprehensive Ocean-Atmosphere Data Set (ICOADS).
\subsubsection{IASI Channel Selection}
\begin{figure}[ht]
  \centering
  \includegraphics[width=0.7\textwidth]{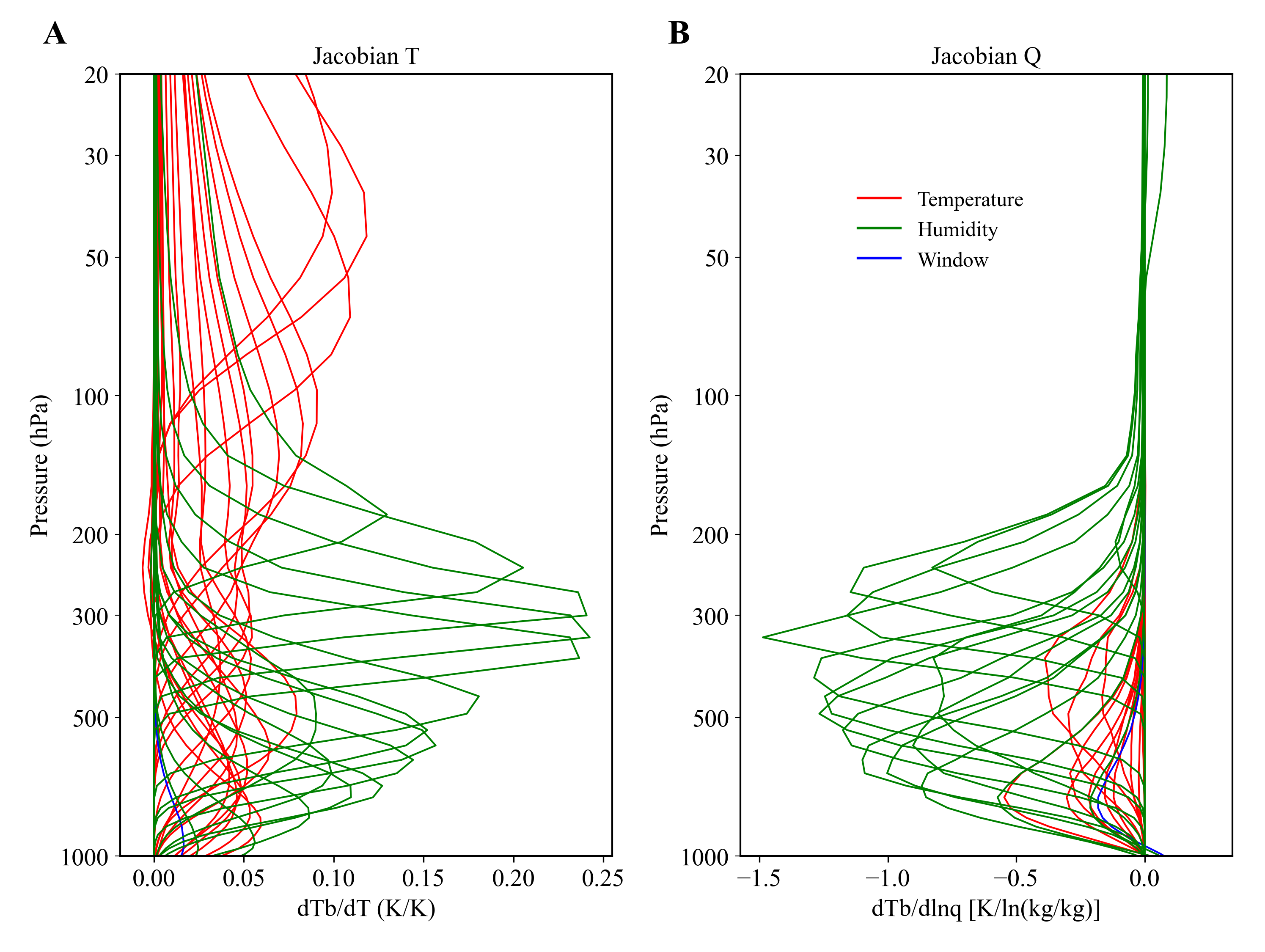}
  \caption{ \textbf{Temperature (A) and humidity (B) Jacobian functions for the selected 38 IASI channels.} A total of 38 channels are included, consisting of 21 channels in the $CO_2$ absorption band (650-770$cm^{-1}$, red lines), 16 channels in the water vapor absorption band (1210-2020$cm^{-1}$, green lines), and 1 atmospheric window channel (943.25$cm^{-1}$, blue line).}
  \label{fig:sup_iasi_channel}
\end{figure}
Directly assimilating all 8,461 IASI channels would impose an immense computational burden and significant information redundancy. To address this, we implemented a refined channel selection strategy. Starting with the 616-channel subset designated for international exchange in Global Data Assimilation System (GDAS), we identified 124 channels that overlap with the 400 highest-information-content channels selected by \cite{coopmann2020update} via information theory. From these overlapping channels, 108 channels were retained, including 85 temperature sounding channels in the $\text{CO}_2$ absorption band ($650\text{--}770 \text{ cm}^{-1}$) and 23 humidity sounding channels in the $\text{H}_2\text{O}$ absorption band ($1210\text{--}2020 \text{ cm}^{-1}$).

Subsequently, to further reduce the number of channels while maintaining vertical sounding capability, we employed an interval sampling strategy based on the peak heights of the temperature and humidity Jacobian. Below the 20 hPa pressure level, only one channel was retained for every 20 hPa pressure increment, reducing the count to 16 temperature channels and 9 humidity channels. Finally, to fill vertical sounding gaps, we strategically added 5 temperature channels, 7 humidity channels, and 1 window channel based on their Jacobian. The final configuration comprises 38 channels, including 21 temperature channels, 16 humidity channels, and 1 window channel. As illustrated in Fig. \ref{fig:sup_iasi_channel}, the temperature and humidity Jacobian distributions of these 38 channels exhibit highly uniform vertical coverage throughout the atmospheric column.
\subsubsection{Quality Control}
We applied a statistical quality control (QC) workflow for all observations before feeding them into FuXiWeather2 model. First, a gross error check was applied to satellite brightness temperatures to constrain observations within a physically plausible range of 50–350 K. 
Second, we implemented a bi-weight-based QC strategy for all observational data types \cite{mapes2003sampling}. The bi-weight method is a robust statistical technique that effectively estimates the mean and standard deviation while reducing distortions caused by extreme outliers \cite{zou2006quality}. This method operates independently of background field comparisons, providing a viable QC solution in scenarios where observation operators are absent.
The bi-weight screening was partitioned into three latitudinal zones (low, $30^{\circ}$S--$30^{\circ}$N; middle, $30^{\circ}$--$60^{\circ}$N/S; and high, $60^{\circ}$--$90^{\circ}$N/S) to account for the latitudinal variations in the global atmospheric mean state. Given the high variance in satellite brightness temperature observations between clear-sky and cloudy conditions, a Z-score threshold of 6 was applied to brightness temperature data, while a more stringent threshold of 4 was enforced for all other observation types.
\subsection{Observation Simulation and Representation}
\subsubsection{Forward Simulation}
\label{sec_forward_sim}
To address the discrepancy in data distribution between the training and operational phases of the assimilation model, we generated a hybrid observation dataset that integrates simulated and real-world observations. Specifically, observations from the MWHS-II, MWTS-III, and GNSS-RO for the period from June 2018 to June 2022 were simulated using physical forward models and ERA5 reanalysis data.

For the MWHS-II and MWTS-III sounders, brightness temperatures were simulated using the RTTOV-SCATT v13 fast radiative transfer model \cite{saunders2018update}. To maintain realistic satellite scanning geometry while optimizing computational efficiency, a proxy replication strategy was implemented. Actual observation coordinates and viewing angles from June 2022 to June 2023 were mapped onto the 2018–2022 period via annual replication (e.g., using the geometric distribution of August 1, 2022, as a template for August 1, 2020). To improve the simulation accuracy of brightness temperatures over land, the surface emissivity was parameterized using the TELSEM2 scheme \cite{aires2011tool}.

For the GNSS-RO data, atmospheric refractivity (N) was derived according to the classic formula by \cite{smith1953constants}:
\begin{equation} \label{eq_state_ref} N = 77.6 \cdot \frac{P}{T}+3.73 \cdot 10^5 \cdot \frac{e}{T^2} \end{equation}
where $P$ is pressure, $T$ is temperature, and $e$ is water vapor pressure. This expression is accurate to 0.5 \% in $N$ for frequencies up to 30 000 MHz and under typical ranges of temperature, pressure, and humidity \cite{smith1953constants}.
To account for the non-uniform spatio-temporal distribution of GNSS-RO data (e.g., the COSMIC-2 constellation provides observations only within the latitude range of ±45 $^{\circ}$), all profiles recorded between June 2022 and June 2023 were aggregated into a 24-hour location lookup table. During each simulation step, a random subset of 500 to 2,500 profiles was sampled from the corresponding hourly table to replicate the global distribution of occultation events.
\subsubsection{Observation Encoding}
\label{sec_obs_encode}
The observation encoding scheme serves as the critical interface bridging irregular, heterogeneous raw observational data and the grid-based deep learning architecture. Our primary objective is to preserve the maximum information content from the original measurements while structuring the data into dense, regular tensors to facilitate efficient representation learning.

Observational data were initially grouped by unique instrument-platform pairs (e.g., separating ATMS data derived from NOAA-20 versus NPP).
Each platform-specific dataset was independently projected onto a regular latitude-longitude grid with a spatial resolution of $0.25^\circ$ ($720 \times 1440$ grid points) using nearest-neighbor interpolation, aligning observations within the window $[T_0, T_0 + 1\text{h})$ to the timestamp $t=T_0$.
Subsequently, these platform-specific tensors were spatially merged to construct the final input. During this integration process, observations from different platforms carrying the same type of instrument may fall into the same grid cell. To resolve these overlaps, we implemented a stochastic selection strategy, randomly retaining a single representative observation for that grid cell. Finally, channel-wise normalization was applied, and grid cells lacking observational coverage were zero-padded to maintain tensor regularity. 

To address the vast disparities in the spatiotemporal densities of various observation types, we implemented two tailored encoding strategies:
\begin{itemize}
\item Continuous Observation Platforms (Temporal Stacking): This branch processes observation types that provide multiple measurements at fixed spatial locations within a short timeframe. This includes satellite sounders providing dense swaths (such as AMSU-A, MHS, ATMS, MWHS-II, MWTS-III, SSMIS, HIRS-4, and IASI), as well as in-situ platforms providing continuous observations at fixed positions (e.g., surface stations and marine platforms). We stack the hourly data within a $T$-hour assimilation window along the temporal dimension, generating a 4D tensor with a shape of ($T \times C \times 720 \times 1440$), where $C$ represents the observational variables. This strategy preserves the integrity of the observations, enabling the model to explicitly extract temporal evolution tendencies from the time-frame dimension.
\item Spatially Dynamic Platforms (Continuous Time Embedding): This strategy is designed for instruments characterized by varying spatial trajectories or asynchronous reporting characteristics, including GNSS-RO and radiosonde observations. For these observations, temporal stacking results in highly sparse tensors dominated by zero-padding. 
To address this, we implemented a continuous time embedding strategy. Instead of discretizing time into separate grid frames, we directly encode the observation time as additional channels to represent the temporal dimension. Specifically, we utilize two continuous time coordinates, $\sin(2\pi \Delta t/W)$ and $\cos(2\pi \Delta t/W)$ (where $W$ is the window size), to encode the offset between the observation time and the start of the assimilation window ($\Delta t$). Using GNSS-RO as an example, this approach transforms a theoretically memory-intensive and highly sparse $(T \times 512 \times 720 \times 1440)$ tensor into a compact representation $(1 \times 514 \times 720 \times 1440)$ covering the entire window. This method significantly reduces memory usage while preserving over 95\% of valid observational information.
\end{itemize}
Finally, for the encoded gridded observations, a mask matrix is created using one-hot encoding to distinguish whether an observation exists at each grid point.
\subsubsection{Spatial dilation strategy}
As the native resolution of most observational instruments is significantly coarser than the target $0.25^\circ$ grid, the encoded tensors often exhibit pronounced spatial sparsity. To mitigate this, we employ a spatial dilation strategy, which propagates valid observational signals to adjacent grid cells, thereby effectively densifying the input tensors and enhancing the model’s capacity to capture sparse features \cite{huang2024diffda}.
Drawing inspiration from the Successive Correction Method (SCM) of utilizing all observations within an influence radius to compute analysis increments \cite{bergthorsson1955numerical,cressman1959operational}, we fill grid points without observations by calculating a weighted average of all observations within a specified influence radius. This procedure is mathematically formulated as:
\begin{equation} y=\frac{\sum_{n=1}^Nw_{n} \cdot y_{n}}{\sum_{n=1}^Nw_{n}} \end{equation} 
where $y$ denotes the filled observation value, $N$ represents the number of observations within the influence radius, $y_n$ signifies the individual observation value within influence radius, and $w_n$ is the weight assigned to the observation $y_n$. The Cressman function is employed to weight the observations within the specified radius \cite{cressman1959operational}:
\begin{equation} w_{n} = \frac{R^2-d_n^2}{R^2+d_n^2}\end{equation} 
where $R$ denotes the influence radius, and $d_n$ represents the distance between the observation $y_n$ and the target filling point. To facilitate GPU acceleration and ensure computational efficiency for high-resolution global grids, we reformulate this filling task as a normalized convolutional operation. Specifically, a convolutional kernel $\mathbf{w} \in \mathbb{R}^{2R \times 2R}$ is derived from the empirical Cressman weighting function:
\begin{equation}
\mathbf{w}_{i,j} = \begin{cases} 
\frac{R^2 - d_{i,j}^2}{R^2 + d_{i,j}^2}, & d_{i,j} < R \\
0, & d_{i,j} \geq R 
\end{cases}
\end{equation}
where $d_{i,j}^2 = (i-R)^2 + (j-R)^2$ represents the squared Euclidean distance from the kernel center to the index $(i, j)$, with an influence radius of $R=10$ and indices $i, j \in \{1, \dots, 2R\}$. 

Let $\mathbf{y} \in \mathbb{R}^{T \times C \times 720 \times 1440}$ represent the sparsely gridded observation tensor and $\mathbf{m} \in \{0, 1\}^{T \times C \times 720 \times 1440}$ represent the corresponding one-hot mask, where $1$ indicates the presence of an observation and $0$ represents a missing value. The densified output $\mathbf{y}_{out}$ is then computed through a normalized convolution process:
\begin{align}
\mathbf{y}_{conv} &= \mathbf{y} * \mathbf{w} \\
\label{eq_mask_conv} \mathbf{m}_{conv} &= \mathbf{m} * \mathbf{w} \\
\label{eq_fill} \mathbf{y}_{out} &= \mathbf{y} \odot \mathbf{m} + \frac{\mathbf{y}_{conv}}{\mathbf{m}_{conv} + \epsilon} \odot (1 - \mathbf{m}) \\
\mathbf{m}_{out} &= \begin{cases} 
1, & \mathbf{m}_{conv} > 0 \\
0, & \mathbf{m}_{conv} = 0 \\
\end{cases}
\end{align}
where $*$ represents the 2D spatial convolution operator, $\odot$ represents the element-wise product, and $\epsilon=10^{-4}$ is a small constant for numerical stability. The resulting $\mathbf{y}_{out}$ ensures that original observational values are strictly preserved where available ($\mathbf{m}=1$), while missing grid cells ($\mathbf{m}=0$) are filled with weighted averages from within the influence radius $R$. $\mathbf{m}_{out}$ signifies the updated mask indicating all grid points that either contain original observations or have been filled by the weighted average of neighbors within the radius $R$. To further distinguish filled points from original observation points, we define a confidence matrix, expressed as:
\begin{equation}
\mathbf{c_{out}} = \begin{cases}
1, & \mathbf{m} > 0 \\
\mathbf{m}_{conv}, & \mathbf{m} = 0 \\
\end{cases} \\
\end{equation}
where $\mathbf{c_{out}} \in \mathbb{R}^{T \times C \times 720 \times 1440}$ denotes the confidence matrix. It assigns a value of 1 to original observation locations, while filled points receive values less than 1. As shown in Equation \ref{eq_mask_conv}, the confidence $\mathbf{c_{out}}$ increases when a filled point is surrounded by more observations. This formulation is reasonable, as a higher density of neighboring observations enhances the reliability of the filled values calculated via Equation \ref{eq_fill}. $\mathbf{c_{out}}$ is concatenated with $\mathbf{y}_{out}$ along the channel dimension to form the final observation matrix. This matrix, together with the corresponding mask matrix $\mathbf{m}_{out}$, is then fed into the assimilation model as input.
\subsection{Problem formulation and model architecture}
We formulate the global weather prediction problem as a unified, cyclic process coupling state estimation and temporal forecasting. FuXiWeather2 is defined by the recursive application of two learnable functional mappings, namely the assimilation module and the forecast module. Both modules operate at a fixed temporal resolution of $\Delta t=6$ hours. The input and output specifications for these modules are provided in Table \ref{tab:sup_module_variables}.

The assimilation module, denoted as $\mathcal{A}(\cdot)$, serves as a learned state estimator. It maps a prior background estimate $\mathbf{x}_{b}^{t}$ and the gridded multi-source observational data $\mathbf{y}^t$ onto an optimal analysis field $\mathbf{x}_a^t$:
\begin{equation} \label{equ_assim} \mathbf{x}_{a}^t = \mathcal{A}(\mathbf{x}_{b}^{t}, \mathbf{y}^t) \end{equation}
where the observational input $\mathbf{y}^t = \{ (\mathbf{v}_1, \mathbf{m}_1), \dots, (\mathbf{v}_K, \mathbf{m}_K) \}^t$ represents the ensemble of observation-mask pairs derived from the spatial dilation strategy. For each observational source $k \in \{1, \dots, K\}$, $\mathbf{v}_k \in \mathbb{R}^{T_k \times C_k \times H \times W}$ denotes the normalized observation tensor, and $\mathbf{m}_k \in \{0, 1\}^{T_k \times C_k \times H \times W}$ is a one-hot mask indicating the spatial availability of the observation, where $T_k$, $C_k$, $H$ and $W$ represent the number of time frames, observed variables, and grid points in latitude and longitude, respectively.

The forecast module, $\mathcal{F}(\cdot)$, operates as a learned autoregressive operator rather than a traditional numerical model. It approximates the temporal evolution of the atmospheric state by directly mapping current states to future states through a single forward pass. To capture temporal evolution, the module predicts the state of the next time step conditioned on the states of two consecutive preceding time steps:
\begin{equation} \label{equ_forecast} \mathbf{x}_{b}^{t + \Delta t} = \mathcal{F}(\mathbf{x}_{a}^{t-\Delta t}, \mathbf{x}_{a}^{t}) \end{equation}
Equations \ref{equ_assim} and \ref{equ_forecast} form a closed-loop cycling system: the forecast output $\mathbf{x}_{b}^{t+\Delta t}$ serves as the prior background for the subsequent assimilation step to generate the next analysis field $\mathbf{x}_{a}^{t+\Delta t}$, implementing a fully differentiable computational graph where both assimilation and forecast modules are learnable.
This allows gradients to propagate through time (Backpropagation Through Time, BPTT) during the unrolled cycling training phase, enabling the model to explicitly minimize cumulative error growth and learn robust error correction from data.

For medium-range weather prediction, the system generates a 10-day trajectory by recursively applying Equation \ref{equ_forecast}. The output of one step serves as the input for the next, accumulating the trajectory in an autoregressive chain: 
\begin{equation} \mathbf{x}_f^{t+k \cdot \Delta t} = \mathcal{F}(\mathbf{x}_f^{t+(k-2) \cdot \Delta t}, \mathbf{x}_f^{t+(k-1) \cdot \Delta t}), \quad k = 1, \dots, 40 \end{equation} 
When k=1, $\mathbf{x}_f^{t-\Delta t} \equiv \mathbf{x}_a^{t-\Delta t}$  and $\mathbf{x}_f^t \equiv \mathbf{x}_a^t$ denote the initial analysis states, as shown in Equation \ref{equ_forecast}. 

In FuXiWeather2, the assimilation module, $\mathcal{A}(\cdot)$, employs a multi-branch U-Net convolutional architecture \cite{sun2025}. The model employs multi-branch, modality-specific encoders to extract features from both the background field and multi-source observational data, which are subsequently integrated through a hierarchical latent fusion mechanism. This allows the model to align raw observational information with the background field directly within the feature space, correcting background errors without the need for an explicit physical observation operator \cite{xu2025}. The forecast component, $\mathcal{F}(\cdot)$ , employs a Swin-Transformer-based architecture \cite{Chen2023}. Unlike previous cascaded approaches \cite{Chen2023, Chen2025}, FuXiWeather2 utilizes a single model to generate the entire 10-day forecast trajectory to avoid the decline in forecast activity typically observed in cascade architectures. \cite{sun2025}.
\subsection{Training and optimization}
\subsubsection{Training objectives}
\label{sec_train_obj}
\begin{figure}[ht]
  \centering
  \includegraphics[width=1.0\textwidth]{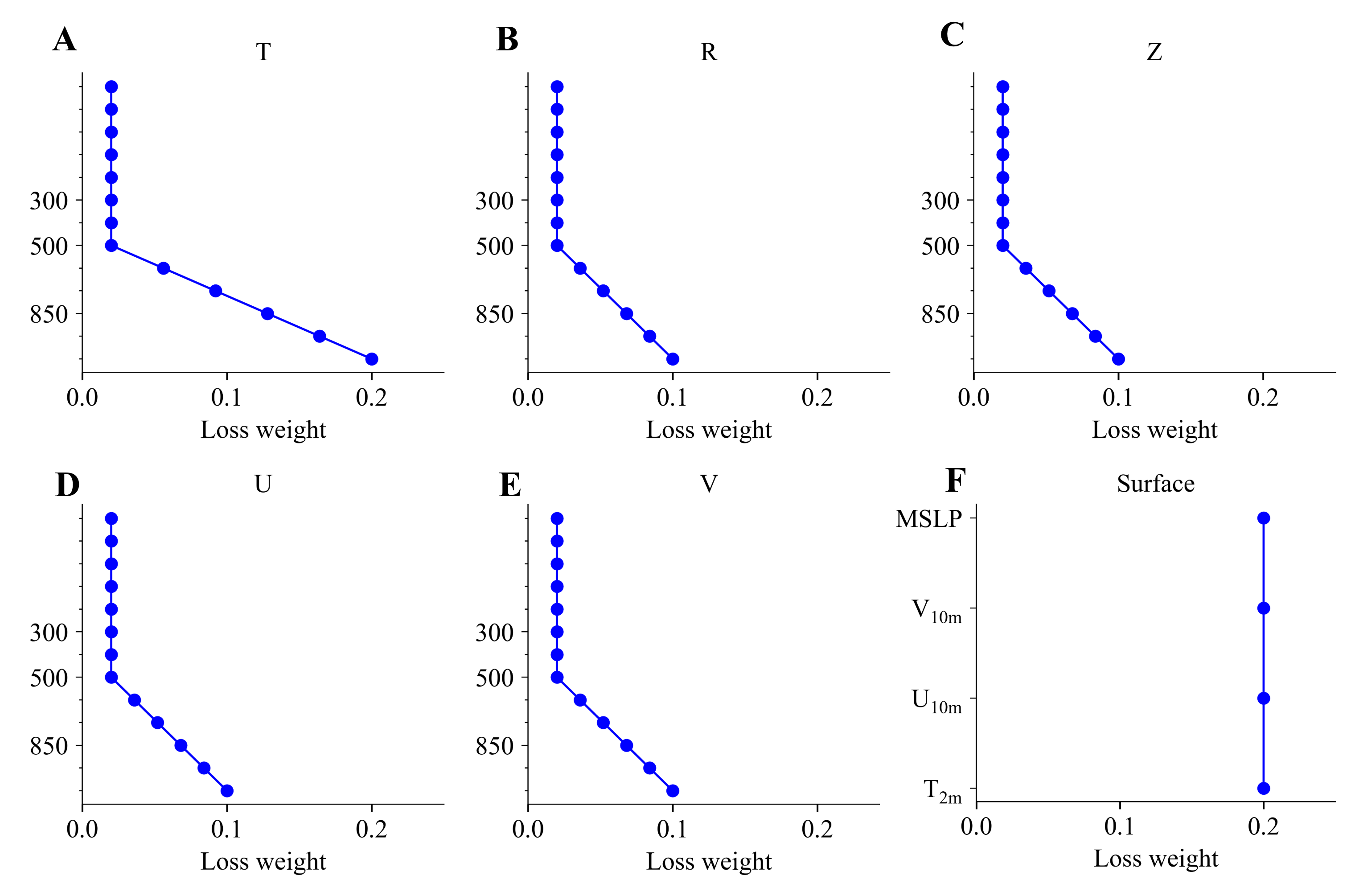}
  \caption{ \textbf{Loss weights for observational supervision.} (A–E) Loss weights for upper-air variables across 13 pressure levels. (F) Loss weights for surface variables. 
  For surface variables, a uniform weight of 0.2 is assigned. For upper-air variables, considering the land–atmosphere interaction, the weights decrease with increasing altitude; specifically, the weights decay linearly with height and remain constant at 0.02 above the 500 hPa level.The maximum weights for relative humidity ($R$), U-wind component ($U$), V-wind component ($V$), and geopotential ($Z$) are set to 0.1, whereas temperature ($T$) is assigned a higher maximum weight of 0.2. 
  This higher weighting for $T$ accounts for its superior spatial continuity, which enhances the spatial representativeness}
  \label{fig:sup_obs_weight}
\end{figure}
To address the challenge where training on a single reanalysis dataset causes the model to inherit both its strengths and its inherent deficiencies, we formulated a composite objective function. This function is designed to leverage the advantages of reanalysis data while compensating for its specific deficiencies. 

First, to ensure that the model output maintains spatiotemporal continuity and adheres to fundamental thermodynamic and dynamical constraints (such as hydrostatic and geostrophic balance), we define the state consistency loss against to the NWP reanalysis fields:
\begin{equation} \label{equ_stat_loss} \mathcal{L}_{state}(\mathbf{x}, \hat{\mathbf{x}}) = \frac{1}{C \cdot H \cdot W} \sum_{c=1}^C \sum_{i=1}^H \sum_{j=1}^W \alpha_i \left| \mathbf{x}_{c,i,j} - \hat{\mathbf{x}}_{c,i,j} \right|\end{equation}
where $\mathbf{x}$ and $\hat{\mathbf{x}}$ denote the model output and ground truth, respectively. In this study, ERA5 is adopted as the reference ground truth. The term $\alpha_i$ represents the latitude-dependent weight:
\begin{equation} \label{eq_lat_weight} \alpha_i = H \cdot \frac{\cos \phi_i}{\sum_{i=1}^H \cos \phi_i}\end{equation}
where $\phi_i$ is the latitude at index $i$. 
This weighting scheme ensures that the error contribution from each grid cell is proportional to its physical surface area \cite{graphcast}.

Although ERA5 provides spatiotemporally continuous and physically consistent supervision signals, it still exhibits inherent simulation deficiencies, particularly in surface variables \cite{sandu2020addressing, de2022coupled}. To address this issue, we introduce an observational consistency loss function during the recursive unrolled training phase, utilizing discrete in-situ observations as direct supervision. This loss function is defined as follows:
\begin{equation} \mathcal{L}_{obs}(\mathbf{x}, \mathbf{y}) = \frac{1}{C} \sum_{c=1}^{C} \lambda_{c} \frac {\sum_{i=1}^{H} \sum_{j=1}^{W} \mathbf{m}_{c,i,j} \cdot \alpha_i \left| \mathbf{x}_{c,i,j} - \mathbf{y}_{c,i,j} \right|}{\sum_{i=1}^{H} \sum_{j=1}^{W} \mathbf{m}_{c,i,j} + 1}  \end{equation} 
where $\mathbf{y}$ represents the gridded observations, and $C$ corresponds to the 69 non-precipitation variables consistent with the output of assimilation module (Table \ref{tab:sup_module_variables}). Among these, the 4 surface variables in $\mathbf{y}$ are a combination of land station observations and marine platform data from ICOADS, while the 65 upper-air variables are derived from radiosondes. $\mathbf{m}$ is the mask matrix using one-hot encoding, where one indicates the presence of an observation and zero indicates its absence.
$\lambda_{c}$ represents the weights for these 69 channels. Due to the significant errors exhibited by ERA5 in surface analysis \cite{sandu2020addressing, de2022coupled}, we assign higher weights to surface observational supervision. Furthermore, a weighting scheme that decays with increasing altitude is implemented for upper-air variables, accounting for the influence of land-atmosphere processes on the near-surface atmosphere (Fig. \ref{fig:sup_obs_weight}).
Finally, we define the total training objective $\mathcal{L}_{joint}$ as: 
\begin{equation}
\label{ref_joint_loss}
\mathcal{L}_{joint}(\mathbf{x}, \hat{\mathbf{x}}, \mathbf{y}) = \mathcal{L}_{state}(\mathbf{x}, \hat{\mathbf{x}}) + \mathcal{L}_{obs}(\mathbf{x}, \mathbf{y}) \end{equation} 
\subsubsection{Overview of training pipeline}
In alignment with advanced operational NWP workflows, FuXiWeather2 consists of two core modules: assimilation module and forecast module. To ensure the stability of the entire training process, we adopt a three-stage curriculum learning strategy, which comprises the following stages: pre-training, recursive unrolled training, and fine-tuning.
\subsubsection{Pre-training}
The assimilation module $\mathcal{A}(\cdot)$ and the forecast module  $\mathcal{F}(\cdot)$ are pre-trained independently to internalize fundamental atmospheric physics before being coupled.

The forecast module, $\mathcal{F}(\cdot)$, is pre-trained on 37 years of ERA5 reanalysis data to learn the temporal evolution of the atmosphere. During this stage, we utilize ERA5 fields as both initial conditions and optimization targets. The model is optimized using a one-step prediction state consistency loss $\mathcal{L}_{state}(\mathbf{x}_{f}^{t+\Delta t}, \hat{\mathbf{x}}^{t+\Delta t})$. By minimizing short-term forecast errors, the forecast model is enabled to generate accurate background fields, thereby ensuring the stability of the subsequent recursive unrolled training. We used the AdamW optimizer with momentum parameters $\beta_1=0.9$ and $\beta_2=0.95$. Training employed a global batch size of 8 distributed across 8 NVIDIA A100 GPUs. The learning rate was initialized at $2.5 \times 10^{-4}$ with a weight decay of 0.1, and the model underwent 40,000 gradient updates. A cosine-annealing schedule was employed to gradually decay the learning rate to zero toward the end of this stage.

The pre-training of $\mathcal{A}(\cdot)$ aims to map background fields and sparse observations to ERA5 reanalysis data. Due to the limited temporal availability of modern satellite observations, this module is trained using data records spanning a five-year period from June 2018 to June 2022, comprising both real-world observations and simulated observations. Crucially, to prevent the module from overfitting to stable background fields, we implement a dynamic background sampling strategy \cite{sun2025}.
During training iterations at time $t$, the background prior $\mathbf{x}_{b}=\mathbf{x}_{f}^{t+k \cdot \Delta t}$ is stochastically sampled from a pre-generated offline forecast dataset with varying lead times ranging from 6 to 180 hours ($k = 1, \dots, 30$). This exposes the module to a broad spectrum of background error characteristics. By optimizing the $\mathcal{L}_{state}(\mathcal{A}(\mathbf{x}_{f}^{t+k \cdot \Delta t}, \mathbf{y}^{t+k \cdot \Delta t}), \hat{\mathbf{x}}^{t+k \cdot \Delta t})$ loss under these diverse input conditions, the assimilation module learns to dynamically adjust the weights between the background prior and the current observational input $\mathbf{y}$. This process ensures that the module possesses strong robustness against background errors encountered during the continuous operational cycling phase. The model was optimized with AdamW ($\beta_1=0.9$, $\beta_2=0.95$) and a weight decay of 0.1. The training was conducted on 16 A100 GPUs distributed across two computing nodes, with a global batch size of 16, and ran for 24,000 gradient updates. We applied a cosine-annealing schedule with warmup: the learning rate increased linearly from 0 to $2.5 \times 10^{-4}$ over the first 5\% of iterations, then decayed to 0 over the remainder.
\subsubsection{Recursive unrolled training}
\label{sec_unrolled_training}
While independent pre-training establishes the fundamental capabilities of each module, it introduces a critical training-inference discrepancy (commonly referred to as exposure bias). During pre-training, the forecast module is initialized with ERA5 reanalysis, and the assimilation module’s background fields are derived from an offline forecast dataset with diverse error characteristics. However, in actual operational cycling, the assimilation and forecast modules must ingest the short-term forecasts and analysis fields generated by one another. Over long-term integration, this inconsistency inevitably leads to the accumulation of errors. To bridge this gap and enhance long-term stability, we transition to a recursive unrolled training stage.

We construct a differentiable cycling graph where the forecast and assimilation modules operate recursively for $N$ steps ($N=4$ in this study). Within each cycle $n \in \{ 1, \ldots, N \}$, the assimilation and forecast modules operate alternately to generate the analysis field $\mathbf{x}_{a}^{t+(n-1) \cdot \Delta t}$ and the background field $\mathbf{x}_{b}^{t+n \cdot \Delta t}$. To better simulate the stable operating state during inference (the stabilization of analysis errors), the system must achieve numerical stability within a limited number of unrolling steps. Initializing the recursive unrolled training with an uninformative background field (e.g., zero-valued field) typically requires an excessive number of unrolling steps to continuously assimilate observations before reaching stability, which significantly increases memory demands. Consequently, we adopt a warm-start strategy by stochastically sampling from the offline forecast dataset to initialize the training. Since these forecast fields already contain atmospheric information, the number of unrolling steps required to reach a steady state is substantially reduced. This approach effectively lowers training costs while ensuring consistency between the training and inference phases.

To rigorously constrain the trajectory evolution, we minimize a cumulative objective over the unrolled steps. We apply the joint loss function (Equation \ref{ref_joint_loss}) to both the analysis and background states: 
\begin{equation} \mathcal{L}_{cycle} = \sum_{n=1}^N[\mathcal{L}_{joint}(\mathbf{x}_{a}^{t+(n-1) \cdot \Delta t}, \hat{\mathbf{x}}^{t+(n-1) \cdot \Delta t},\mathbf{y}^{t+(n-1) \cdot \Delta t})+\mathcal{L}_{joint}(\mathbf{x}_{b}^{t+n \cdot \Delta t}, \hat{\mathbf{x}}^{t+n \cdot \Delta t},\mathbf{y}^{t+n \cdot \Delta t})] \end{equation} 
We optimized the model with AdamW ($\beta_1=0.9$, $\beta_2=0.95$) and a weight decay of 0.1. The training was conducted on 4 compute nodes, each with 8 GPUs (32 GPUs total), using a global batch size of 16. The training ran for 4,000 gradient updates. We adopted the same warmup plus cosine-decay learning-rate schedule as in the assimilation pre-training stage, but with a maximum learning rate of $10^{-4}$.

\subsubsection{Fine-tuning}
Upon completion of the recursive unrolled training stage, FuXiWeather2 is capable of generating stable, high-quality analysis fields. To enhance the long-range forecasting capability of the forecast module, we adopt an autoregressive training regime and curriculum training schedule \cite{graphcast,Chen2023}. The training loss function is formulated as follows:
\begin{equation} \mathcal{L}_{multi} = \frac{1}{N}\sum_{n=1}^N \mathcal{L}_{state}(\mathbf{x}_{f}^{t+n \cdot \Delta t}, \hat{\mathbf{x}}^{t+n \cdot \Delta t}) \end{equation} 
where $N$ denotes the number of autoregressive steps, which is increased from 1 to 20 in this study. The model was optimized with AdamW ($\beta_1=0.9$, $\beta_2=0.95$) and a weight decay of 0.1. This stage was run on one compute nodes with 8 A100 GPUs, using a global batch size of 4. We employed a curriculum learning schedule by gradually increasing the forecast horizon from 1 to 20 steps, performing 500 gradient updates for each increment. A fixed learning rate of $1e^{-7}$ was maintained throughout this training stage.
\begin{figure}[ht]
  \centering
  \includegraphics[width=1.0\textwidth]{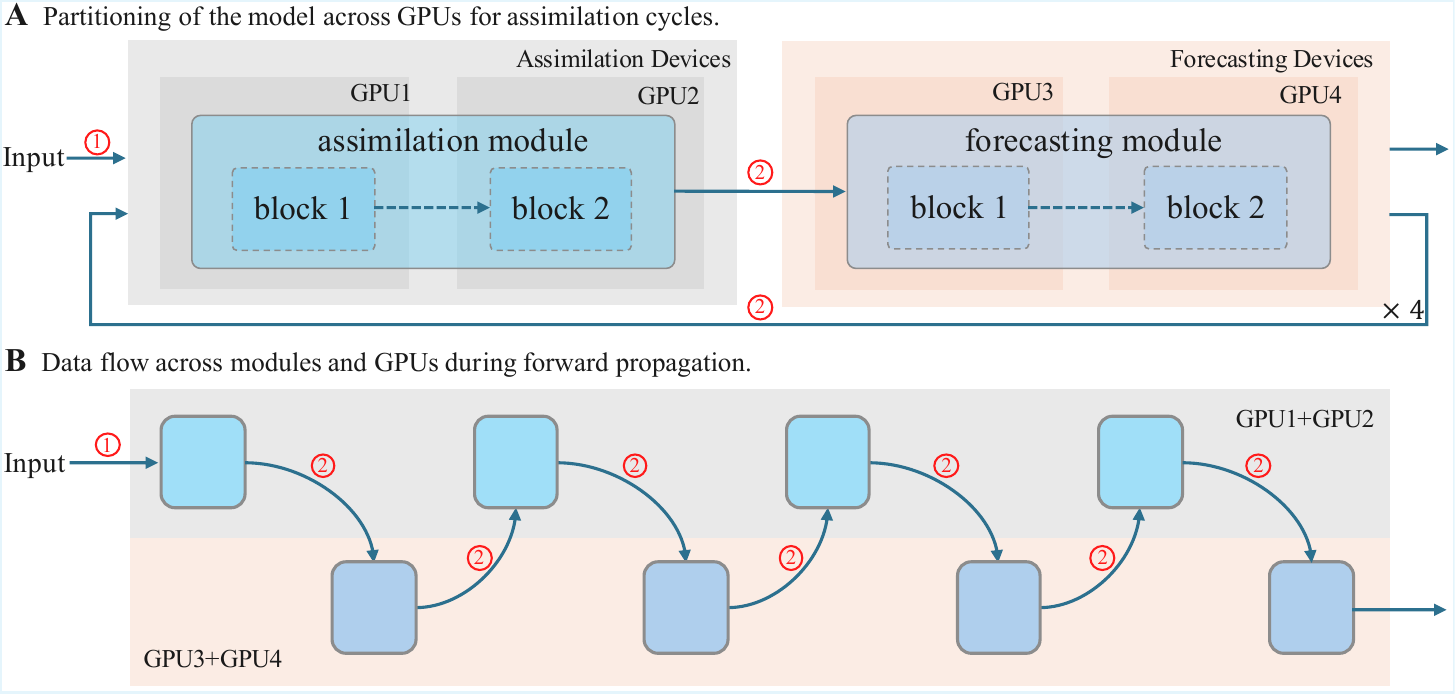}
  \caption{ \textbf{Schematic of the pipeline parallelism strategy for recursive unrolled training.} (A) Partitioning of the model across GPUs for assimilation cycles. (B) Data flow across modules and GPUs during forward propagation. FuXiWeather2 distributes the tightly coupled assimilation and forecasting modules across four GPUs. GPU1 and GPU2 serve as assimilation devices, while GPU3 and GPU4 act as forecasting devices. Each module is further divided into two sequential blocks. The $\times 4$ symbol denotes the continuous execution of four assimilation cycles, where the forecast output cyclically feeds back into the assimilation module. }
  \label{fig:sup_pipline}
\end{figure}
\subsubsection{Computational Implementation}
\label{sec_compu_imple}
During the recursive unrolled training phase, FuXiWeather2 integrates tightly coupled assimilation and forecasting modules, whose joint training poses substantial challenges on existing computational infrastructure. Although our computing nodes are equipped with A100 GPUs with 80 GB of memory, this capacity remains insufficient for end-to-end training of the full system.

The excessive memory demand arises from three primary sources. First, each training iteration requires loading high-resolution background fields alongside massive volumes of multi-source observational data. Second, in addition to model parameters, the training process must retain gradients, optimizer states, and intermediate activations to support backpropagation. For data-driven models operating at high spatial resolution, the memory footprint of these auxiliary states substantially exceeds that of the model parameters themselves. Third, and most critically, the memory requirement grows linearly with the number of assimilation cycles, rendering single-GPU training infeasible.
Although standard memory-optimization techniques such as activation checkpointing and mixed-precision training were applied, they proved insufficient to meet these extreme memory demands. Ultimately, it was the implementation of a customized pipeline parallelism strategy that successfully enabled the end-to-end joint training of the coupled system.

As illustrated in Fig. \ref{fig:sup_pipline}, the entire computational workload is distributed across multiple GPUs, meticulously partitioning both the functional modules and their internal layers. Specifically, the computational resources are divided into two primary groups: assimilation devices (GPU1 and GPU2) and forecasting devices (GPU3 and GPU4). Rather than placing an entire module on a single device, we further partition the internal structures of both models. The assimilation module is split into two sequential blocks, with block 1 residing on GPU1 and block 2 on GPU2. Similarly, the forecasting module is partitioned across GPU3 and GPU4. During the forward pass, data flow is transferred sequentially across GPU boundaries: from GPU1 to GPU2 to compute the analysis fields, and subsequently to GPU3 and GPU4 to generate the background forecasts.

Crucially, to support the multi-cycle assimilation process without exceeding memory limits, the pipeline is designed to accommodate cyclic data flows. Upon completing one cycle, the output from the forecasting module on GPU4 is sequentially fed back to the assimilation module on GPU1 to initiate the next cycle. 

By distributing the model parameters, optimizer states, and most importantly, the massive intermediate activations across four separate devices, this pipeline parallelism approach effectively prevents out-of-memory (OOM) errors. It guarantees that the peak memory usage on any single GPU remains within the 80 GB hardware limit, making the joint training of high-resolution, multi-cycle DA and forecasting models computationally feasible and efficient.
\subsection{Evaluation metrics}
To evaluate the analysis fields, we compare both the analysis and the analysis-driven short-term forecast fields against in-situ observations (radiosondes, land stations and marine platforms data). The corresponding root mean square error (RMSE), mean bias error (MBE) and standard deviation (STD) values are:
\begin{equation} \label{eq_insitu_rmse} \textrm{RMSE}_{\textrm{in-situ}} = \sqrt{\frac{1}{N} \sum_{n=1}^{N} [H(\mathbf{x}) - \mathbf{y}]^2 } \end{equation}
\begin{equation} \label{eq_insitu_mbe} \textrm{MBE}_{\textrm{in-situ}} = \frac{1}{N} \sum_{n=1}^{N} [H(\mathbf{x}) - \mathbf{y}] \end{equation}
\begin{equation} \label{eq_insitu_std} \textrm{STD}_{\textrm{in-situ}} = \sqrt{\frac{1}{N} \sum_{n=1}^{N} [H(\mathbf{x}) - \mathbf{y} - \textrm{MBE}_{\textrm{in-situ}}]^2 } \end{equation}
Where $x$ is the analysis or short-term forecast field, $y$ is the observations, $N$ is the number of observations, and $H$ is the observation operator. For evaluations based on in-situ observations, $H$ specifically refers to nearest-neighbor interpolation, which is used to achieve the spatial matching between the analysis or short-term forecast fields and the in-situ observational data.

While in-situ observations are restricted to specific geographic locations, we further incorporated GNSS-RO refractivity data, which offer a more uniform global distribution, for the evaluation of analysis fields. The RMSE, MBE and STD can be expressed as:
\begin{equation} \label{eq_gnss_rmse} \textrm{RMSE}_{\textrm{GNSS-RO}} = \sqrt{\frac{1}{N} \sum_{n=1}^{N} [\frac{H(\mathbf{x}) - \mathbf{y}}{H(\mathbf{x})}]^2 } \end{equation}
\begin{equation} \label{eq_gnss_rmse} \textrm{MBE}_{\textrm{GNSS-RO}} = \frac{1}{N} \sum_{n=1}^{N} [\frac{H(\mathbf{x}) - \mathbf{y}}{H(\mathbf{x})}] \end{equation}
\begin{equation} \label{eq_gnss_rmse} \textrm{STD}_{\textrm{GNSS-RO}} = \sqrt{\frac{1}{N} \sum_{n=1}^{N} [\frac{H(\mathbf{x}) - \mathbf{y}}{H(\mathbf{x})} - \textrm{MBE}_{\textrm{GNSS-RO}}]^2 } \end{equation}
Here, $H$ specifically represents the combination of the conversion from atmospheric state to refractivity (Equation \ref{eq_state_ref}) and nearest-neighbor interpolation. For GNSS-RO data, the spatial matching must account for both horizontal and vertical dimensions. In the vertical direction, we require that the geopotential difference between the observation point and the meteorological field must be less than 1,000 $m^2 \cdot s^{-2}$. The $H(x)$ in the denominator of Equation \ref{eq_gnss_rmse} is used for normalization, as refractivity values decrease exponentially with height; this is a widely adopted approach in refractivity evaluation \cite{anthes2008cosmic,anthes2018estimating,gilpin2018reducing,schreiner2020cosmic,xu2025quality}.

For global forecast evaluation, we employ the latitude-weighted RMSE and Anomaly Correlation Coefficient (ACC):
\begin{equation} \label{RMSE_fc_equation} \textrm{RMSE} = \sqrt{\frac{1}{H \cdot W} \sum_{i=1}^{H}\sum_{j=1}^{W} \alpha_i \left( \mathbf{x}_{i,j} - \hat{\mathbf{x}}_{i,j} \right)^{2}} \end{equation} 
\begin{equation} \label{ACC_equation} \textrm{ACC} = \frac{\sum_{i}^{H}\sum_{j}^{W} \alpha_i (\mathbf{x}_{i,j} - \mathbf{m}_{i,j}) (\hat{\mathbf{x}}_{i,j} - \mathbf{m}_{i,j})} {\sqrt{ \sum_{i}^{H}\sum_{j}^{W} \alpha_i (\mathbf{x}_{i,j} - \mathbf{m}_{i,j})^2 \cdot \sum_{i}^{H}\sum_{j}^{W} \alpha_i(\hat{\mathbf{x}}_{i,j} - \mathbf{m}_{i,j})^2}}\end{equation}
where $m$ denotes the climatological mean, calculated using ERA5 reanalysis data spanning the period 1993–2016, and $\alpha_i$ represents the latitude-dependent weight described in Equation \ref{eq_lat_weight}.

The normalized RMSE difference between experiment A and experiment B is calculated as $(\textrm{RMSE}_{\textrm{A}}-\textrm{RMSE}_{\textrm{B}})/\textrm{RMSE}_{\textrm{B}}$, and the normalized ACC difference is calculated as $(\textrm{ACC}_{\textrm{A}}-\textrm{ACC}_{\textrm{B}})/\textrm{ACC}_{\textrm{B}}$. Additionally, we used the paired normalized RMSE difference to perform the t-test of significance \cite{geer2016significance}.
\section{Results}
\subsection{Global weather analysis}
\begin{figure}[ht]
  \centering
  \includegraphics[width=1\textwidth]{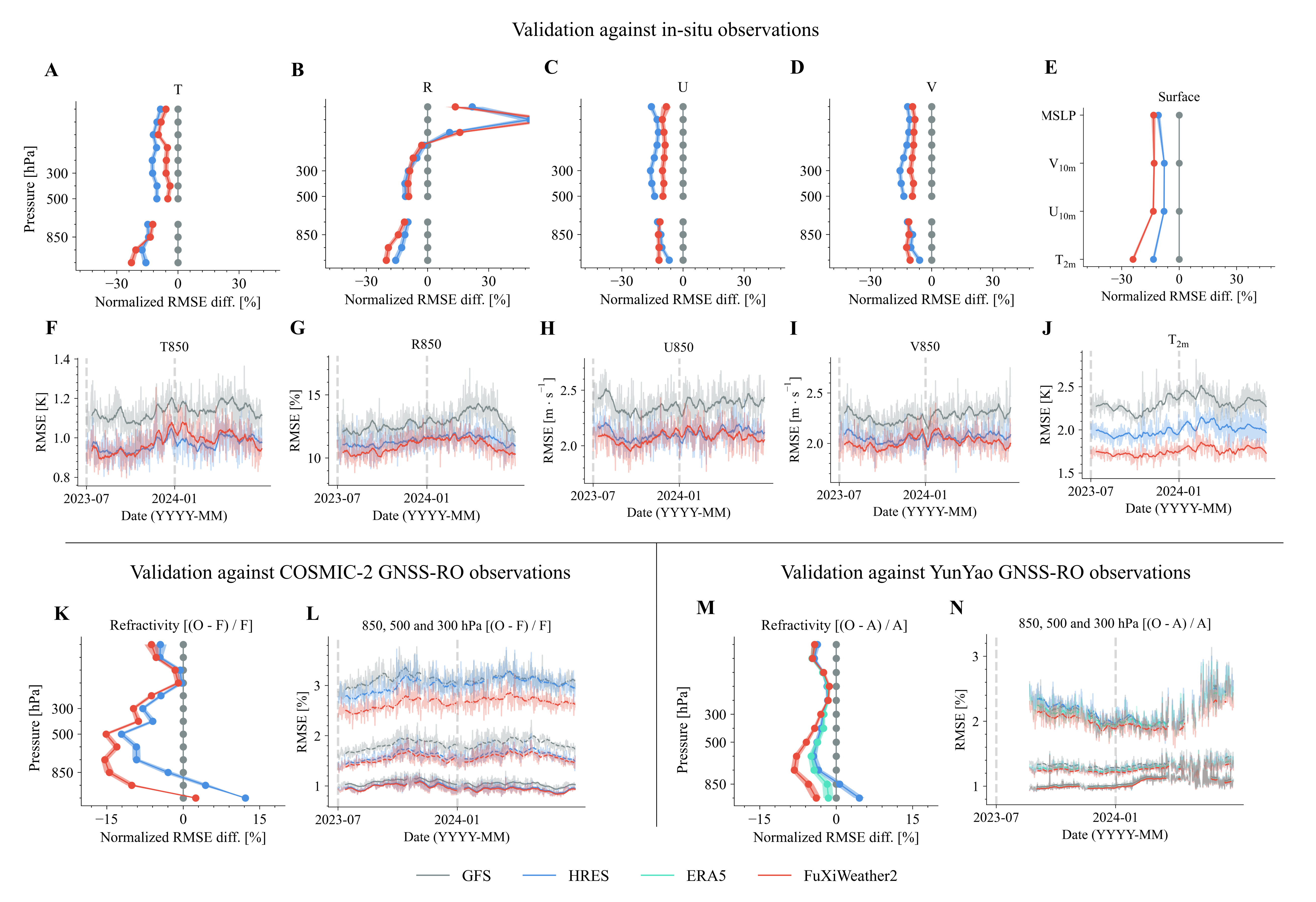}
  \caption{\textbf{Observation-based verification for global analysis.} (A–J) Time-averaged normalized root mean square error (RMSE) differences (A-E) and RMSE time series (F-J) of 12-hour forecasts using in-situ observations as ground truth. (K, L) Time-averaged normalized RMSE differences (K) and RMSE time series (L) of 12-hour forecasts using COSMIC-2 refractivity data as ground truth. (M, N) Time-averaged normalized RMSE differences (M) and RMSE time series (N) of analysis fields using YunYao refractivity data as ground truth. Gray, blue, green, and red lines represent GFS, HRES, ERA5, and FuXiWeather2, respectively. The evaluation spans a one-year testing period at 00:00 and 12:00 UTC.
  In panels L and N, dash-dotted, dashed, and solid lines denote the 850, 500, and 300 hPa pressure levels, respectively. In panels A-E, K and M, 13 pressure levels are displayed, with shaded areas represent the three times the 95\% confidence intervals of the t-test.
  Evaluation is conducted on 12-hour forecast fields (A-L) for observations already assimilated (radiosonde, land station, marine platform, and COSMIC-2 GNSS-RO) to ensure independence. For YunYao GNSS-RO data, which remain unassimilated by all products, the evaluation is performed directly on the (re)analysis fields (M and N).}
  \label{fig:analysis_obs}
\end{figure}
\begin{figure}[htbp]
\centering
\includegraphics[width=1\textwidth]{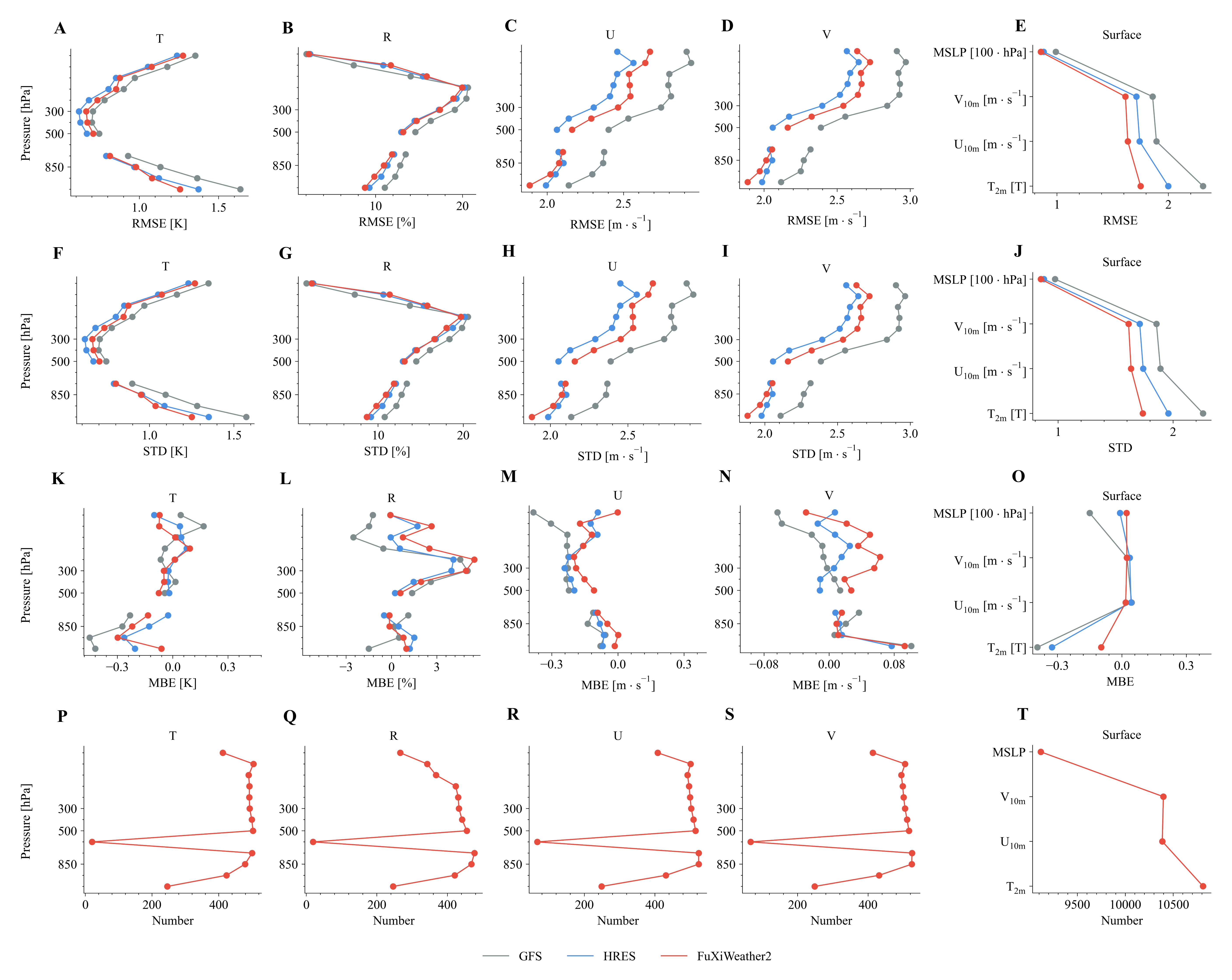}
\caption{\textbf{Global analysis verification using in-situ observations as ground truth} (A–O) Time-averaged root mean square error (RMSE, A-E), standard deviation (STD, F-J), and mean bias error (MBE, K-O) of 12-hour forecasts. (P–T) Time-averaged number of observations used for validation. Radiosonde observations are used as the ground truth for upper-air variables across 13 pressure levels, while the combination of land station and marine platform observations serves as the ground truth for surface variables. Gray, blue, and red lines represent GFS, HRES, and FuXiWeather2, respectively. The evaluation spans a one-year testing period at 00:00 and 12:00 UTC and is performed on the 12-hour forecast fields, as the in-situ observations have already been assimilated into all the analysis products.}
\label{fig:sup_ana_insitu}
\end{figure}
\begin{figure}[htbp]
\centering
\includegraphics[width=1.0\textwidth]{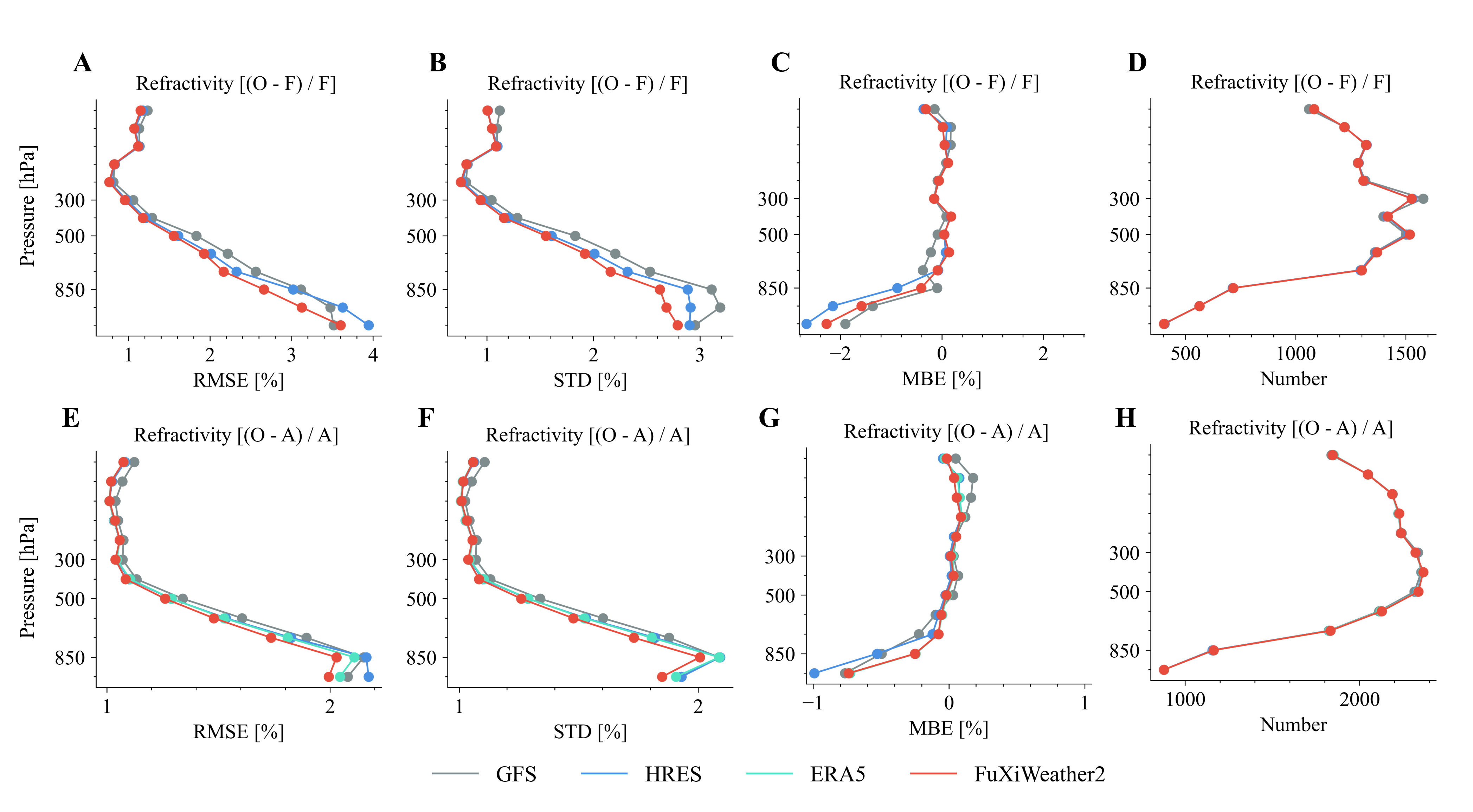}
\caption{\textbf{Global analysis verification using GNSS-RO observations as ground truth} (A–C) Time-averaged root mean square error (RMSE, A), standard deviation (STD, B), and mean bias error (MBE, C) of 12-hour forecasts using COSMIC-2 refractivity data as ground truth. (D) Time-averaged number of COSMIC-2 observations used for validation. (E–G) Time-averaged RMSE (E), STD (F), and MBE (G) of (re)analysis fields using YunYao refractivity data as ground truth. (H) Time-averaged number of YunYao observations used for validation. Gray, blue, green, and red lines represent GFS, HRES, ERA5, and FuXiWeather2, respectively. The evaluation spans a one-year testing period at 00:00 and 12:00 UTC across 13 pressure levels. For COSMIC-2 GNSS-RO data, the evaluation is conducted on the 12-hour forecast fields because these observations have already been assimilated into all the analysis products (A-D). For YunYao GNSS-RO data that have not been assimilated by any product, the evaluation is performed directly on the (re)analysis fields (E-H).}
\label{fig:sup_ana_gnss}
\end{figure}
\begin{figure}[ht]
\centering
\includegraphics[width=1.0\textwidth]{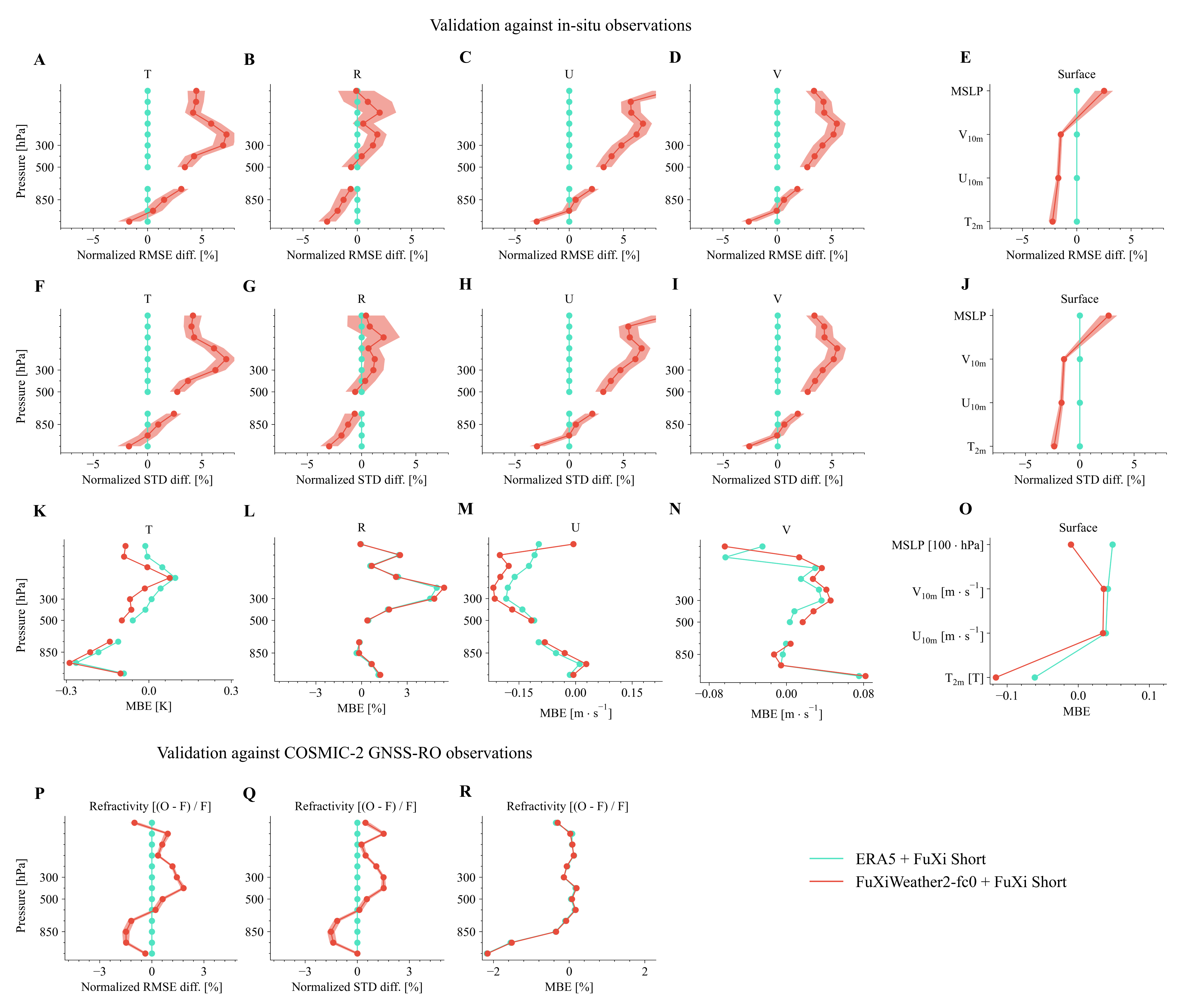}
\caption{\textbf{Observation-based verification of 12-hour forecasts generated by FuXi-Short, initialized with ERA5 and FuXiWeather2 analysis fields (FuXiWeather2-fc0).} (A–O) Time-averaged normalized root mean square error (RMSE) differences (A-E), normalized standard deviation (STD) differences (F-J) and mean bias error (MBE, K-O) using in-situ observations as ground truth. (P-R) Time-averaged normalized RMSE differences (P), normalized STD differences (Q) and MBE (R) using COSMIC-2 refractivity data as ground truth. Green and red lines represent ERA5 and FuXiWeather2, respectively. The evaluation spans a one-year testing period, including forecasts initialized at 00:00 and 12:00 UTC. In panels A-J, P and Q, the shaded areas represent the 95\% confidence intervals of the t-test}
\label{fig:sup_ana_obs_fuxishort}
\end{figure}
\begin{figure}[ht]
\centering
\includegraphics[width=1.0\textwidth]{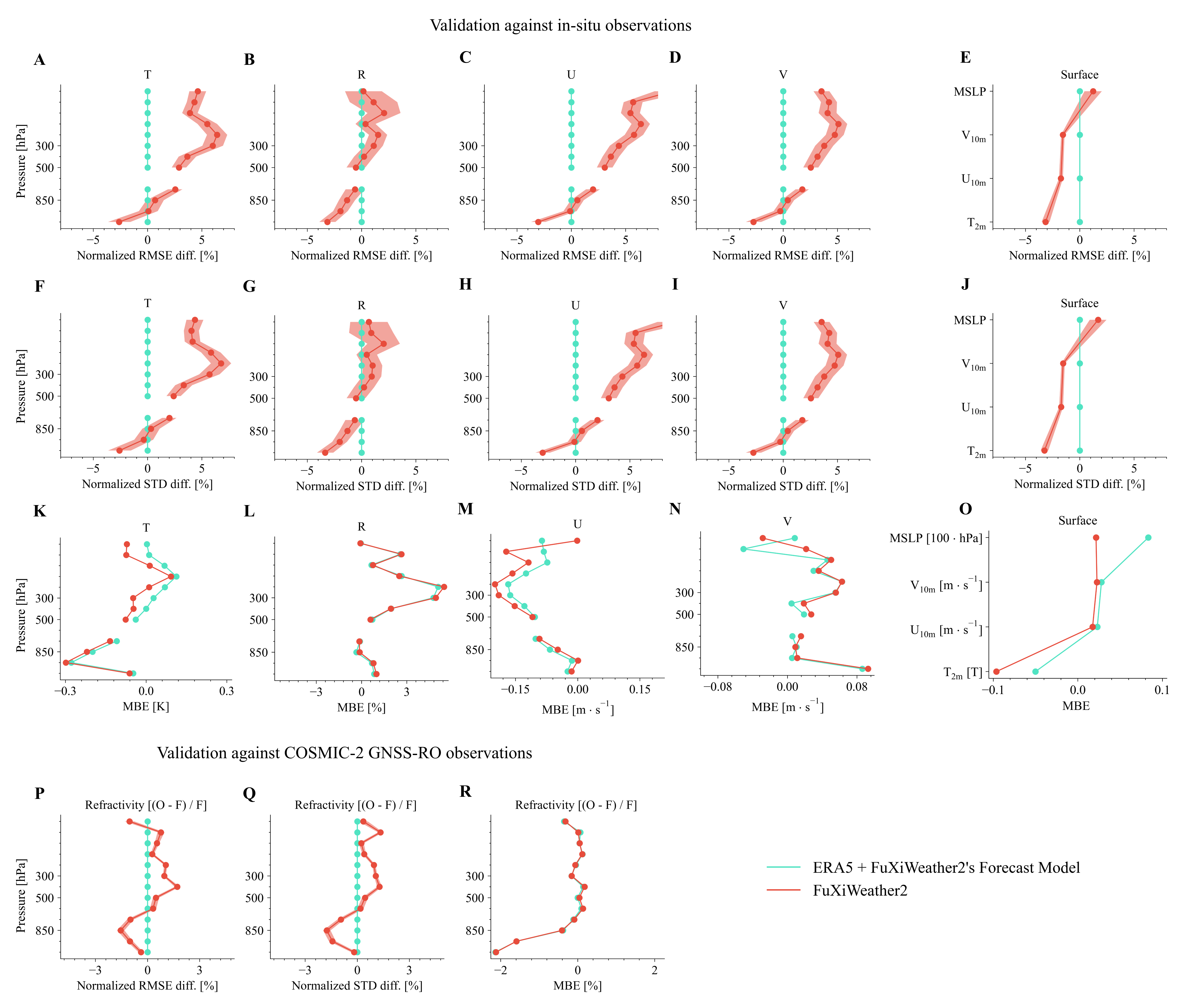}
\caption{\textbf{Observation-based verification of 12-hour forecasts generated by FuXiWeather2's forecast model, initialized with ERA5 and FuXiWeather2 analysis fields (FuXiWeather2-fc0).} (A–O) Time-averaged normalized root mean square error (RMSE) differences (A-E), normalized standard deviation (STD) differences (F-J) and mean bias error (MBE, K-O) using in-situ observations as ground truth. (P-R) Time-averaged normalized RMSE differences (P), normalized STD differences (Q) and MBE (R) using COSMIC-2 refractivity data as ground truth. Green and red lines represent ERA5 and FuXiWeather2, respectively. The evaluation spans a one-year testing period, including forecasts initialized at 00:00 and 12:00 UTC. In panels A-J, P and Q, the shaded areas represent the 95\% confidence intervals of the t-test}
\label{fig:sup_ana_obs_fw2fcst}
\end{figure}
\begin{figure}[htbp]
\centering
\includegraphics[width=1\textwidth]{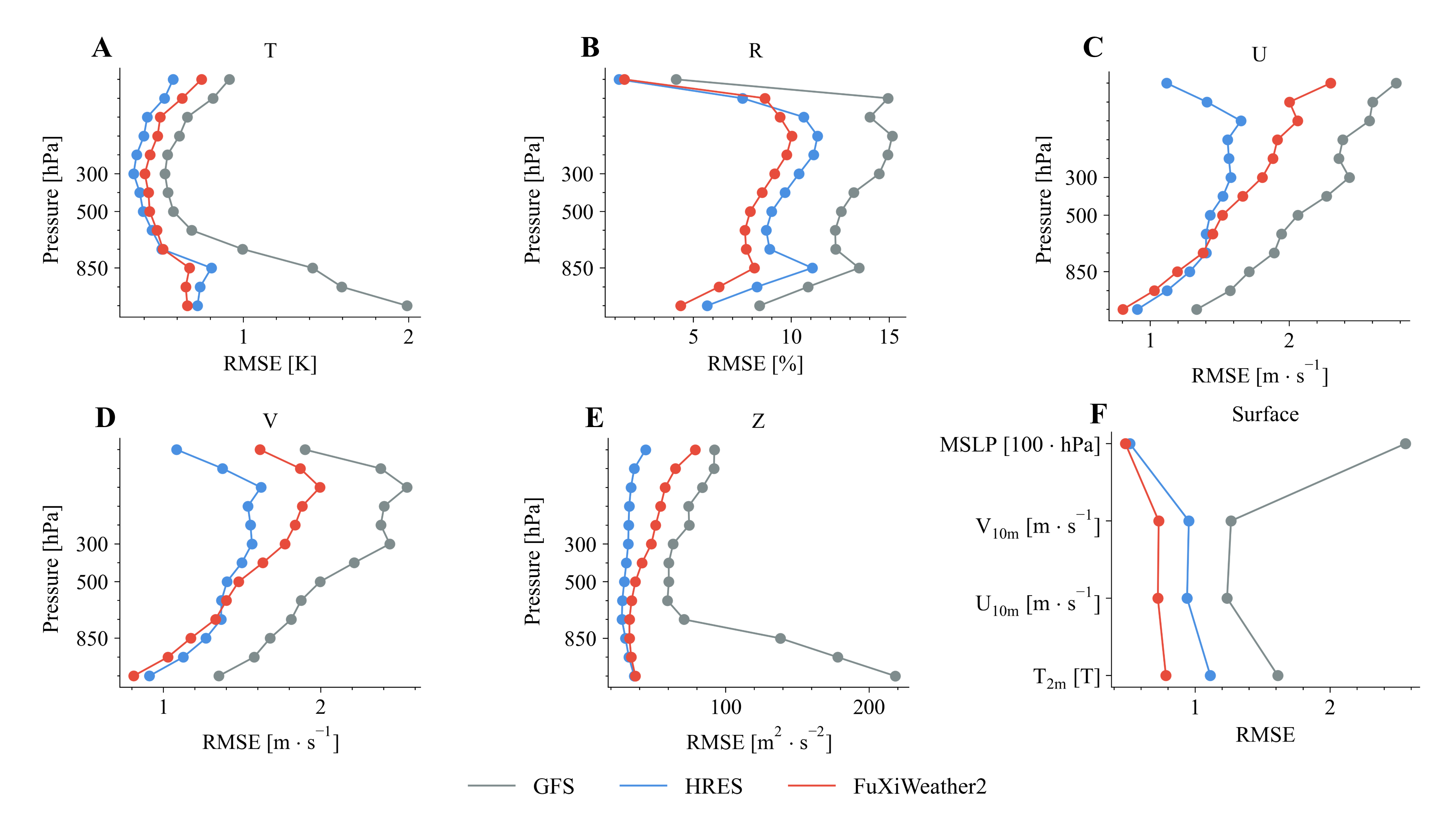}
\caption{\textbf{Global analysis verification using ERA5 as ground truth} (A–F) Time-averaged latitude-weighted root mean square error (RMSE) for five upper-air variables (A-E) across 13 pressure levels and four surface variables (F). Gray, blue, and red lines represent GFS, HRES, and FuXiWeather2, respectively. The evaluation spans a one-year testing period at 00:00 and 12:00 UTC.}
\label{fig:sup_ana_era5}
\end{figure}
\begin{figure}[htbp]
  \centering
\includegraphics[width=1\textwidth]{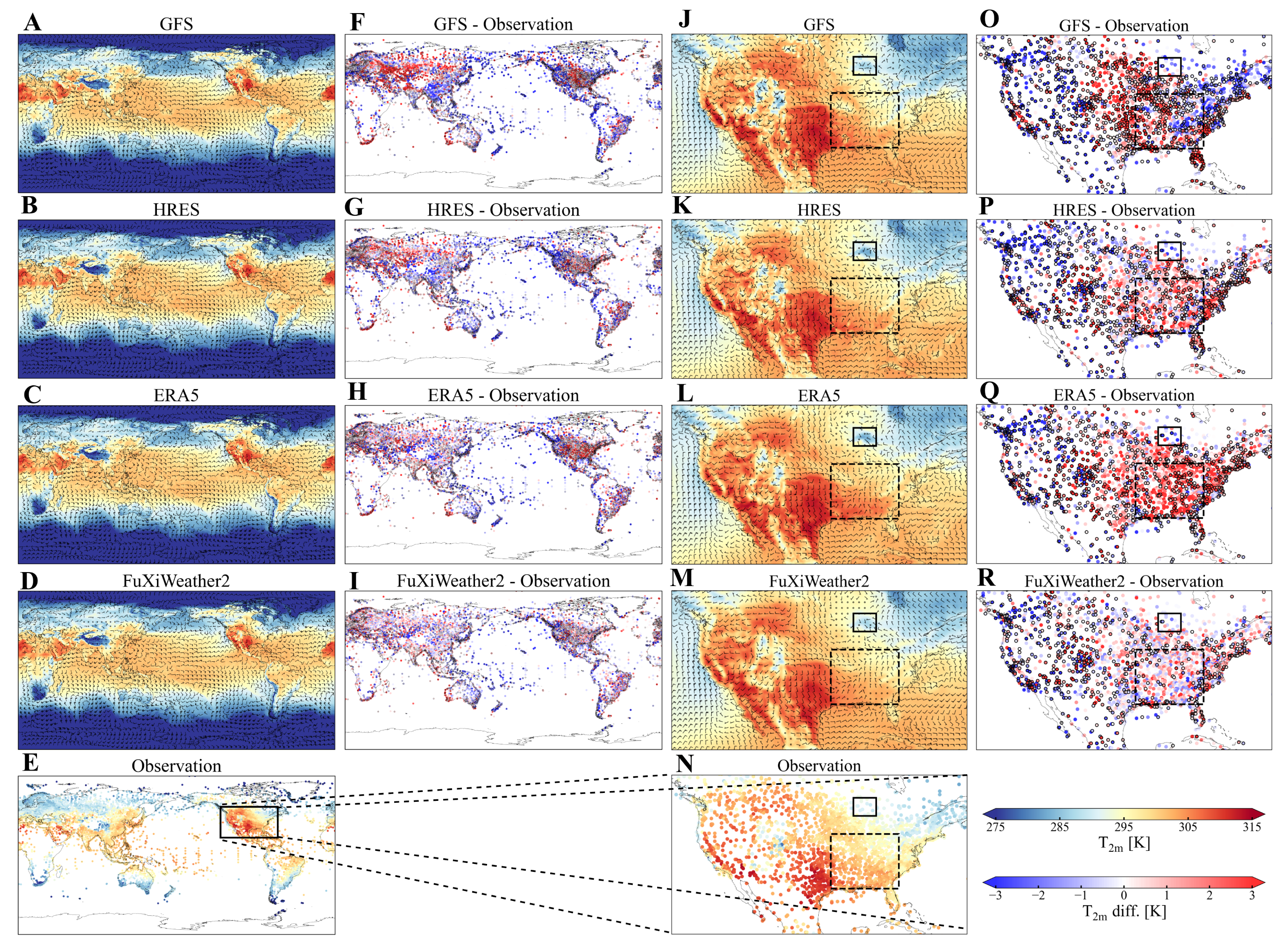}
  \caption{ \textbf{Comparison of surface variables at 00 UTC, 1 August 2023.} (A-D) Global distributions of 2-meter temperature ($T_{2m}$) and wind fields for GFS, HRES, ERA5, and FuXiWeather2. (E) Global distributions of $T_{2m}$ observations. (F-I) Global distributions of the differences between the analysis and observations for GFS, HRES, ERA5, and FuXiWeather2. (J-N) $T_{2m}$ and wind fields distribution over United State for GFS, HRES, ERA5, and FuXiWeather2. (N) $T_{2m}$ observation distribution over United State. (O-R) Differences between the analysis and observations over United State for GFS, HRES, ERA5, and FuXiWeather2. In panels A-D and J-M, within the wind vector plots, a long barb represents 4 m/s, a short barb represents 2 m/s, and a pennant indicates 20 m/s. In panels F-I, and O-R, dots marked with black outlines indicate wind speed biases exceeding 2 m/s between the analysis field and the station observations.}
  \label{fig:analysis_obs_map}
\end{figure}
For global analysis fields, we evaluated the performance of FuXiWeather2 using in-situ and Global Navigation Satellite System (GNSS) radio occultation (RO) observations as ground truth, comparing the results with ERA5 reanalysis as well as two leading deterministic operational NWP systems: the ECMWF HRES and the NCEP GFS. Upper-air variables ($R$,$T$,$U$,$V$) were verified against radiosonde and GNSS-RO data, while surface variables ($T_{2m}$, $U_{10m}$, $V_{10m}$, and $MSLP$) were verified using land station observations and marine platform data from the International Comprehensive Ocean-Atmosphere Data Set (ICOADS). 
Since radiosondes, land stations, marine platforms, and COSMIC-2 GNSS-RO observations had already been assimilated into the evaluated analysis products, these evaluations were conducted on 12-hour forecast fields. The commercial YunYao GNSS-RO data \cite{fu2021introduction,xu2025quality} were not assimilated into any of the evaluated products, nor were they included in the training or testing phases of our model. As strictly independent third-party data, they serve as a benchmark for the direct evaluation of the analysis fields.

Fig. \ref{fig:analysis_obs}A–J present the comparison of 12-hour forecast fields from FuXiWeather2, HRES, and GFS against in-situ observations at 00:00 and 12:00 UTC. FuXiWeather2 significantly outperforms GFS across nearly all evaluated variables and shows superior performance over HRES in lower tropospheric and surface variables (Fig. \ref{fig:analysis_obs}A–E), as evidenced by lower root mean square error (RMSE) values. In terms of time series, FuXiWeather2 consistently maintains an edge over GFS, and its performance for $R850$, $U850$, $V850$, and $T_{2m}$ surpasses that of HRES for the majority of the evaluation period (Fig. \ref{fig:analysis_obs}F–J). Compared to HRES, FuXiWeather2 exhibited larger errors at higher atmospheric levels. This discrepancy was possibly caused by the assimilation of fewer upper-air observations compared to HRES, such as atmospheric motion vectors (AMVs, \cite{schmetz1993operational,nieman1997fully}). Moreover, FuXiWeather2's short-term forecasts exhibit smaller RMSE values in evaluations based on COSMIC-2 GNSS-RO data, particularly within the lower atmosphere. Notably, in the evaluation using independent YunYao data, FuXiWeather2 analysis field exhibits lower errors below the 300 hPa level compared to the ERA5 reanalysis, HRES, and GFS analysis fields, providing robust evidence of its high precision in the lower atmosphere. 

Fig. \ref{fig:sup_ana_insitu} presents a more detailed comparison using in-situ observations as truth, including the RMSE, STD, and MBE. Since the in-situ observations have already been assimilated by GFS, HRES, and FuXiWeather2, the comparison is conducted using 12-hour forecasts. FuXiWeather2 exhibits smaller RMSE and STD values in the lower troposphere and for surface variables. While almost all variables for GFS, HRES, and FuXiWeather2 show relatively consistent MBE, the bias of FuXiWeather2 for $T_{2m}$ is significantly smaller than those of GFS and HRES. 
Fig. \ref{fig:sup_ana_gnss} displays the comparison results using GNSS-RO refractivity as the ground truth. Since COSMIC-2 observations have been assimilated by all (re)analysis systems, the evaluation against this dataset is conducted on 12-hour forecast fields to ensure the independence of the assessment. In contrast, YunYao GNSS-RO observations have not been assimilated by any system; thus, they serve as completely independent third-party observations for the direct evaluation of the analysis fields. As shown in the figure, FuXiWeather2 exhibits significantly smaller RMSE and STD below the 300 hPa level in comparisons against both COSMIC-2 and YunYao data. All (re)analyses and 12-hour forecasts exhibit similar bias patterns: a negative bias is observed at and below the 850 hPa level, while the bias remains near zero above 850 hPa, which is consistent with previous studies \cite{schreiner2020cosmic,xu2025quality}.

To avoid the impact of forecast model discrepancies on the comparative results, Figs \ref{fig:sup_ana_obs_fuxishort} and \ref{fig:sup_ana_obs_fw2fcst} provides a comparison of 12-hour forecasts using ERA5 and FuXiWeather2 analysis fields as inputs for a fixed forecast model. Two forecast models are employed: the original FuXi-Short model (Fig. \ref{fig:sup_ana_obs_fuxishort}) and the forecast model within FuXiWeather2 (Fig. \ref{fig:sup_ana_obs_fw2fcst}). The evaluation is conducted using in-situ and COSMIC-2 GNSS-RO observations as truth. In the evaluation against in-situ observations, FuXiWeather2 demonstrates smaller RMSE and STD for both surface variables and atmospheric variables below 850 hPa. The bias characteristics align with the findings in Fig \ref{fig:sup_ana_insitu}, where ERA5 and FuXiWeather2 exhibit consistent bias patterns across most variables. Furthermore, when validated using COSMIC-2 as the ground truth, FuXiWeather2 maintains superior RMSE and STD performance below the 600 hPa level. These consistent results across multiple reference datasets further consolidate the advantages of the FuXiWeather2 analysis fields in representing the lower troposphere and surface states.

Consistent with previous studies using ERA5 as the ground truth \cite{allen2025,sun2025,gupta2026healda}, Fig \ref{fig:sup_ana_era5} displays the comparison of analysis fields. As illustrated, FuXiWeather2's analysis fields exhibit a significantly lower RMSE compared to earlier research \cite{sun2025,gupta2026healda}. Relative to HRES, the performance gains of FuXiWeather2 in $T$, $U$, and $V$ are primarily concentrated in the lower troposphere. The $R$ variable shows a consistent advantage across nearly all vertical levels, while Z remains comparable to HRES below 850 hPa. For surface variables, FuXiWeather2 consistently achieves a lower RMSE than HRES.

Fig. \ref{fig:analysis_obs_map} presents an example of the surface analysis field comparison. As shown in Fig. \ref{fig:analysis_obs_map}A-D, the global spatial distributions of $T_{2m}$ and wind fields generated by FuXiWeather2 are highly consistent with NWP analysis (ERA5, HRES, and GFS), demonstrating its capability to accurately capture large-scale global horizontal thermodynamic and dynamic structures. Furthermore, compared to observations, FuXiWeather2 exhibits significantly lower $T_{2m}$ biases on a global scale, particularly in western Russia and the United States (Fig. \ref{fig:analysis_obs_map}F-I). A detailed examination of the United States region further validates the advantages of FuXiWeather2: while ERA5 and HRES show notable warm biases in the southern United States (dashed boxes) and cold biases in the north (solid boxes), the biases in FuXiWeather2 are markedly smaller (Fig. \ref{fig:analysis_obs_map}O-R). Additionally, the dots marked with black outlines in the figure indicate wind speed biases exceeding 2 m/s between the (re)analysis fields and station observations. As illustrated, FuXiWeather2 has significantly fewer sites with wind speed biases greater than 2 m/s compared to other (re)analysis fields (Fig. \ref{fig:analysis_obs_map}F-I and O-R), verifying the superior accuracy of the FuXiWeather2 surface wind analysis.
\subsection{Global weather forecasting}
For global weather forecasting, we compared the 10-day forecasts from FuXiWeather2 with those of HRES and GFS, initialized at 00 and 12 UTC with a 6-hour forecast step. All models were verified using their own analysis fields as the ground truth. Evaluation metrics include the globally-averaged and latitude-weighted RMSE and anomaly correlation coefficient (ACC) for surface variables and selected key atmospheric pressure-level variables.
\begin{figure}[tbp]
  \centering
  \includegraphics[width=1.0\textwidth]{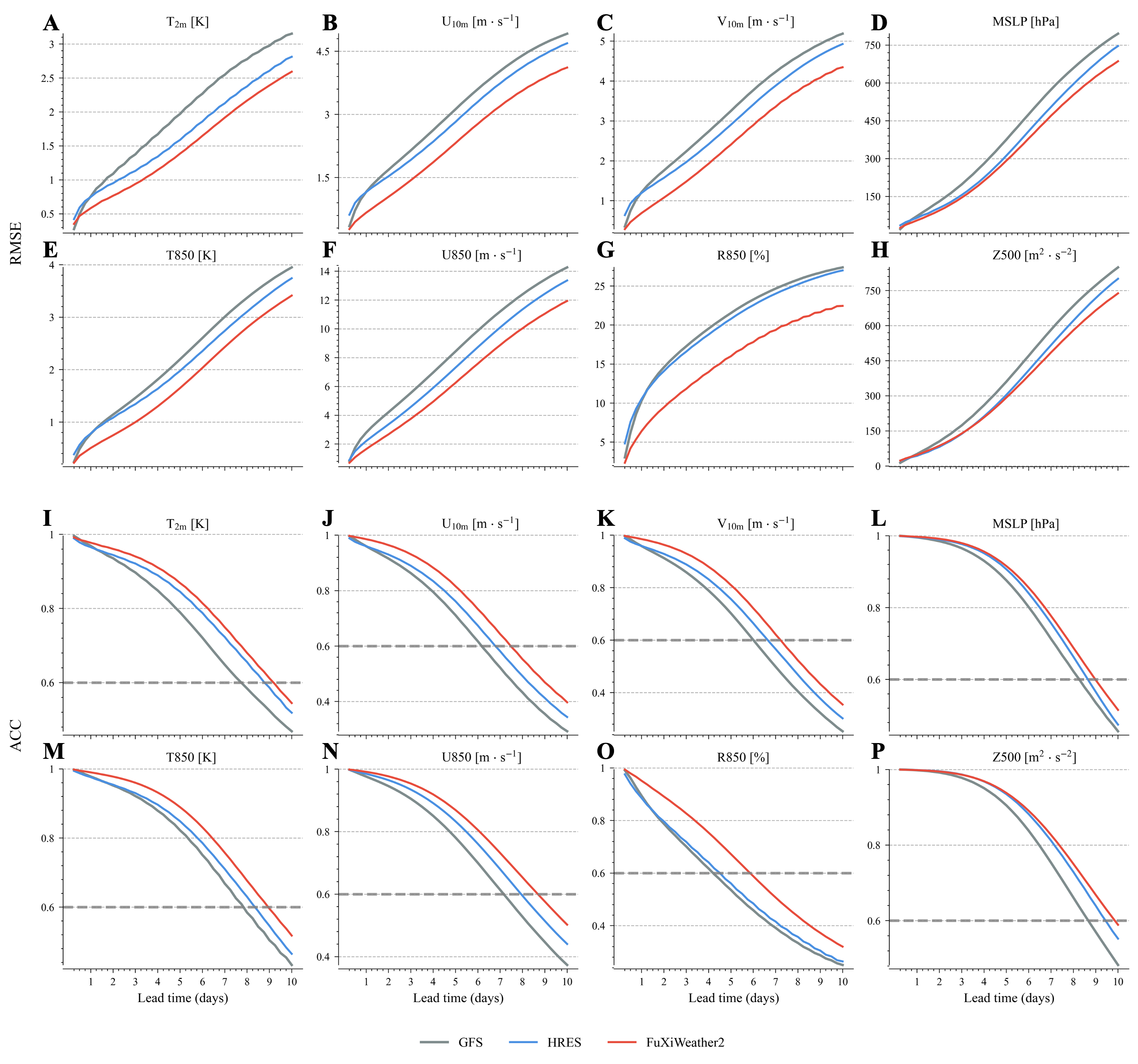}
  \caption{ \textbf{Comparison of 10-day forecast performance.} (A–P) Globally-averaged and latitude-weighted root mean square error (RMSE, A-H) and anomaly correlation coefficient (ACC, I-P) for selected variables. 
  The figure displays four surface variables: 2-meter temperature ($T_{2m}$, A and I), 10-meter U-wind component ($U_{10m}$, B and J), 10-meter V-wind component ($V_{10m}$, C and K), and mean sea-level pressure ($MSLP$, D and L), as well as four key upper-air atmospheric variables: temperature at 850 hPa ($T850$, E and M), relative humidity ($R$) at 850 hPa ($R850$, F and N), U-wind component at 850 hPa ($U850$, G and O), and geopotential at 850 hPa ($Z500$, H and P).
  Gray, blue, and red lines represent GFS, HRES, and FuXiWeather2, respectively. The evaluation spans a one-year testing period, including forecasts initialized at 00:00 and 12:00 UTC with a 6-hour forecast step. All forecasts are verified using their own analysis fields as the ground truth.}
  \label{fig:fcst_rmse_self}
\end{figure}
\begin{figure}[ht]
  \centering
  \includegraphics[width=1\textwidth]{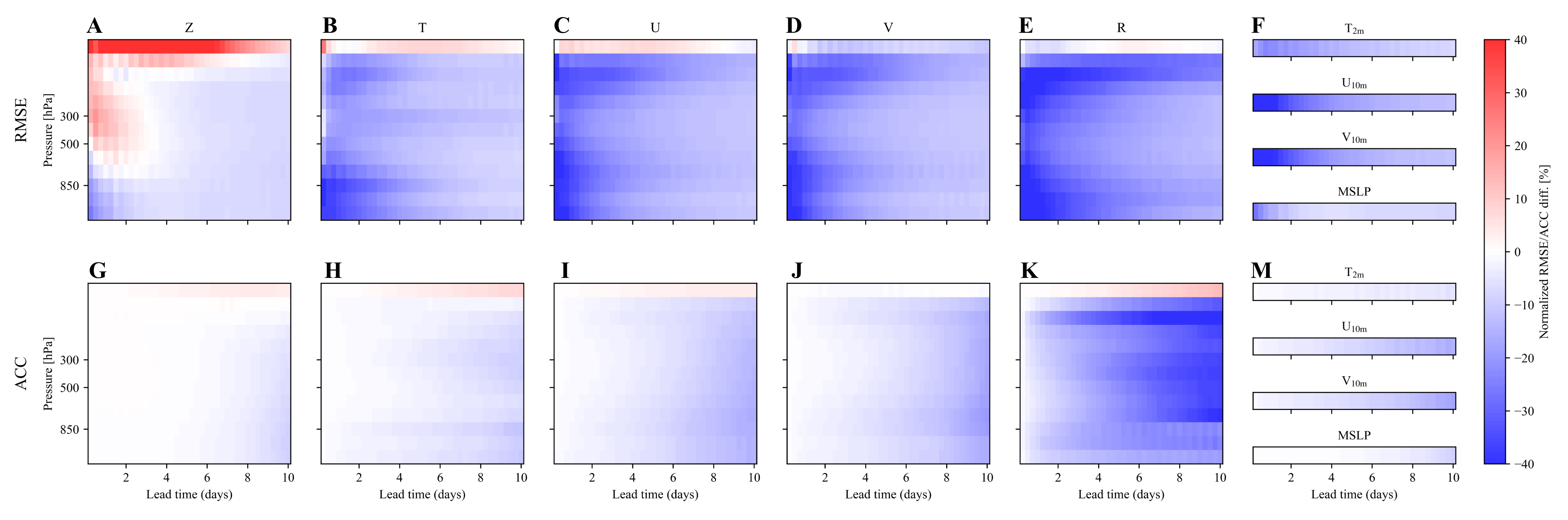}
  \caption{ \textbf{Comparison of 10-day forecast performance between FuXiWeather2 and HRES, verified using their own analysis fields as ground truth.} (A-F) Globally averaged, latitude-averaged root mean square error (RMSE) change rates of FuXiWeather2 relative to HRES for upper-air variables (A-E) across 13 pressure levels and surface variables (F). (G-M) Globally averaged, latitude-averaged anomaly correlation coefficient (ACC) change rates of FuXiWeather2 relative to HRES for upper-air variables (G-K) and surface variables (M). The evaluation spans a one-year testing period, including forecasts initialized at 00:00 and 12:00 UTC with a 6-hour forecast step}
  \label{fig:sup_fcst_all}
\end{figure}
\begin{figure}[ht]
  \centering
  \includegraphics[width=1\textwidth]{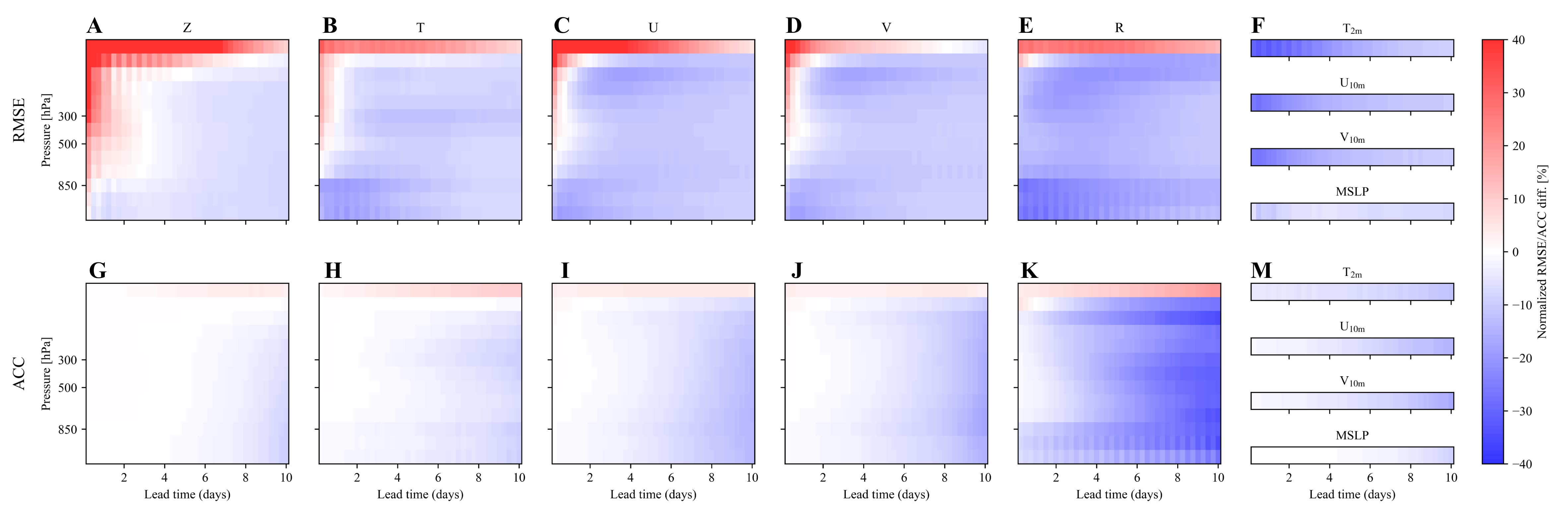}
  \caption{ \textbf{Comparison of 10-day forecast performance between FuXiWeather2 and HRES, verified using ERA5 as ground truth.} (A-F) Globally averaged and latitude-averaged root mean square error (RMSE) change rates of FuXiWeather2 relative to HRES for upper-air variables (A-E) across 13 pressure levels and surface variables (F). (G-M) Globally averaged and latitude-averaged anomaly correlation coefficient (ACC) change rates of FuXiWeather2 relative to HRES for upper-air variables (G-K) and surface variables (M). The evaluation spans a one-year testing period, including forecasts initialized at 00:00 and 12:00 UTC with a 6-hour forecast step}
  \label{fig:sup_fcst_all_era5}
\end{figure}
\begin{figure}[ht]
  \centering
  \includegraphics[width=0.8\textwidth]{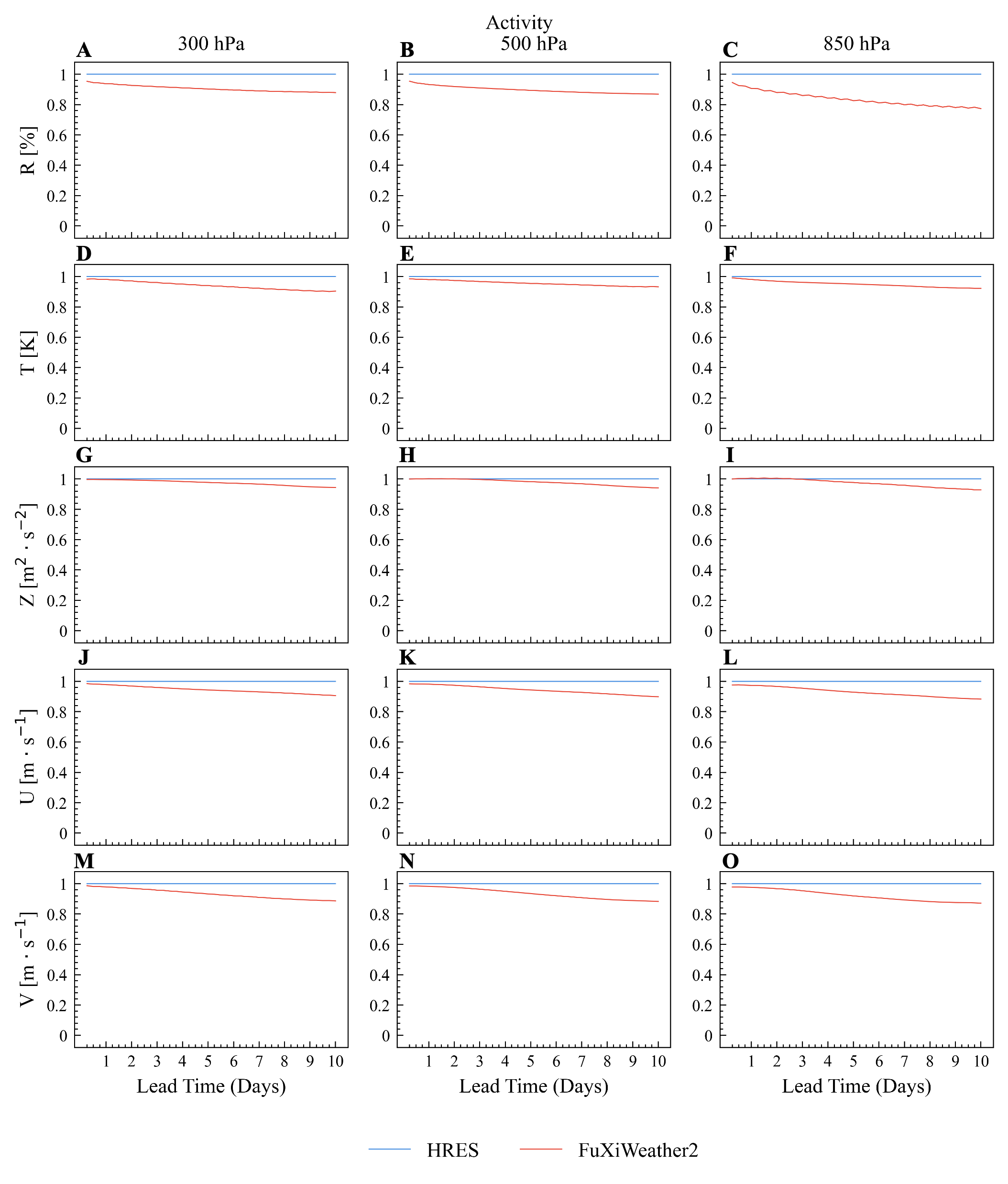}
  \caption{ \textbf{Comparison of std-based forecast activity for upper-air variables.} The figure presents the globally-averaged and latitude-weighted forecast activity for forecasts generated by the FuXiWeather2 and HRES over a 10-day forecasts, covering five upper-air variables across 3 pressure levels. The evaluation spans a one-year testing period, including forecasts initialized at 00:00 and 12:00 UTC with a 6-hour forecast step. Forecast activity is scaled relative to HRES activity.}
  \label{fig:sup_fcst_activity_upper}
\end{figure}
\begin{figure}[ht]
  \centering
  \includegraphics[width=0.8\textwidth]{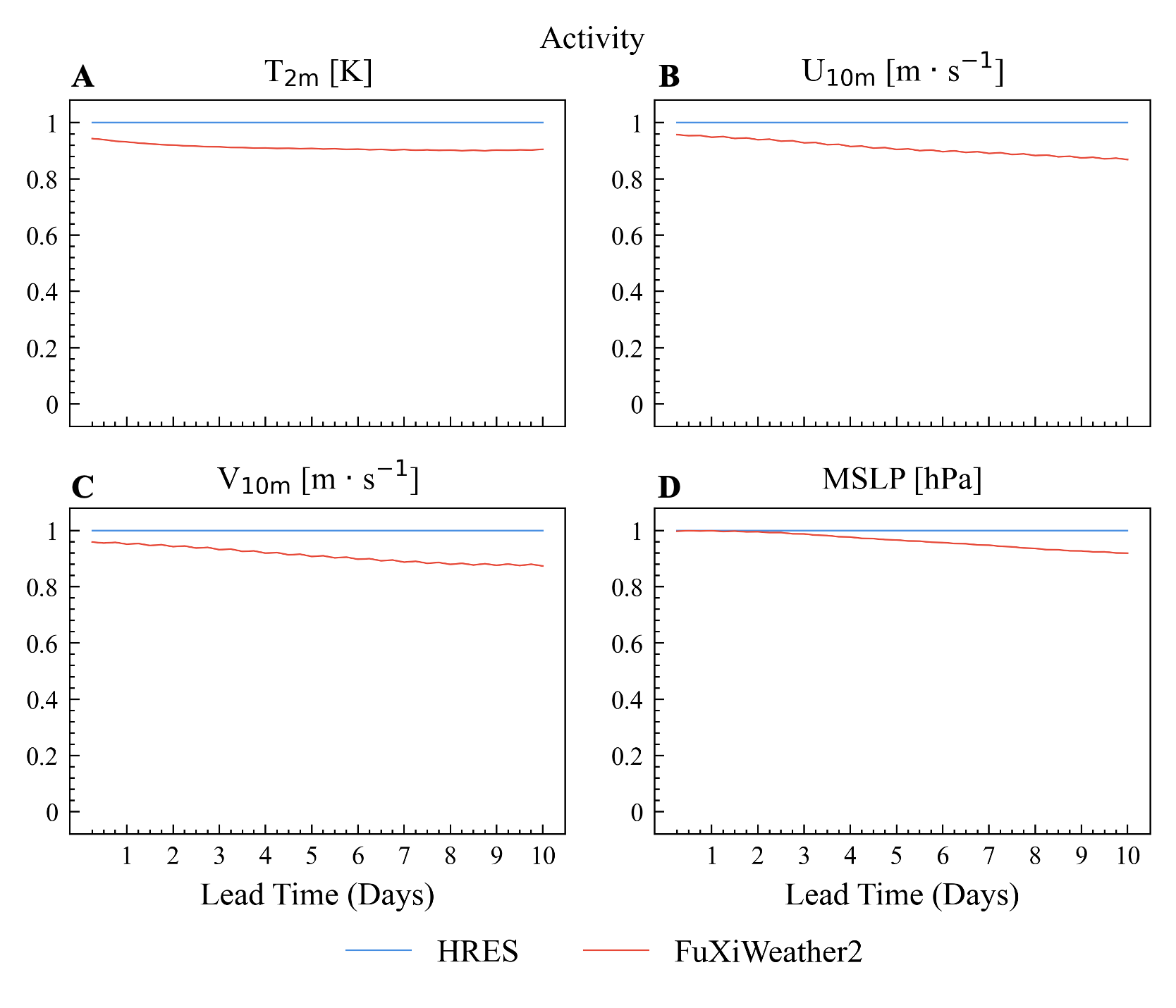}
  \caption{ \textbf{Comparison of std-based forecast activity for surface variables.} The figure presents the globally-averaged and latitude-weighted forecast activity for forecasts generated by the FuXiWeather2 and HRES over a 10-day forecasts, covering four surface variables. The evaluation spans a one-year testing period, including forecasts initialized at 00:00 and 12:00 UTC with a 6-hour forecast step. Forecast activity is scaled relative to HRES activity.}
  \label{fig:sup_fcst_activity_surf}
\end{figure}
Fig. \ref{fig:fcst_rmse_self}A–H present the comparison of globally-averaged and latitude-weighted RMSE. Regarding surface variables, FuXiWeather2 significantly outperforms GFS and HRES, exhibiting lower RMSE. For upper-air variables, FuXiWeather2 demonstrates reduced forecast errors for $T850$, $U850$, and $R850$; for $Z500$, its performance is comparable to HRES during the first three days of the forecast and shows a clear advantage thereafter. Notably, the short-term forecast skill of FuXiWeather2 shows a marked enhancement over previous end-to-end models \cite{allen2025,sun2025}, a benefit attributed to the improved quality of the analysis fields. The ACC evaluation results are highly consistent with those of the RMSE (Fig. \ref{fig:fcst_rmse_self}I–P). Specifically, the effective forecast lead time for Z500 (defined as ACC $\ge$ 0.6) reaches 9.75 days for FuXiWeather2, surpassing the 9.25 days achieved by the contemporaneous HRES.

Fig. \ref{fig:sup_fcst_all} presents the comparison results between FuXiWeather2 and HRES across all evaluated variables. As shown in the figure, the normalized difference in globally-averaged and latitude-weighted RMSE and ACC reveals that FuXiWeather2 consistently outperforms HRES for all surface variables ($T_{2m}$, $U_{10m}$, $V_{10m}$, and $MSLP$) throughout the 10-day forecast period. For upper-air variables, FuXiWeather2 exhibits superior performance for in predicting $T$, $R$, $U$, and $V$ across nearly all lead times, with the exception of the 50 hPa level. While FuXiWeather2 shows higher RMSE for $Z$ at upper levels during the initial lead days, it successfully surpasses HRES in the subsequent forecast period. Overall, across the combinations of all 69 variables and 40 forecast lead times, FuXiWeather2 outperforms HRES in 91.34\% of the RMSE metrics and 88.99\% of the ACC metrics. 

The comparison using ERA5 as the ground truth is shown in Fig. \ref{fig:sup_fcst_all_era5}, where FuXiWeather2 continues to demonstrate a clear advantage for surface variables and the lower troposphere. For upper-air levels above approximately 850 hPa, although FuXiWeather2 exhibits larger RMSE during the initial lead time, it outperforms HRES for most variables beyond 24 hours. Ultimately, FuXiWeather2 achieves superior performance over HRES in 85.40\% of cases for RMSE and 84.75\% for ACC.

Figs. \ref{fig:sup_fcst_map_r}-\ref{fig:sup_fcst_map_v} display the spatial distribution of the RMSE for HRES and FuXiWeather2 forecasts, along with the spatial distribution of the RMSE differences between the two systems. These values are calculated relative to their own analysis fields without latitude weighting.
The maps cover five upper-air variables (${R}$, $T$, ${Z}$, ${U}$, and ${V}$) at three pressure levels (300, 500, and 850 hPa) as well as three surface variables. The evaluated forecast lead times include 1, 3, and 6 days. 
With increasing lead times, both FuXiWeather2 and HRES exhibit a trend of increasing RMSE across all variables, which is particularly pronounced in the mid-latitude regions.

The spatial distribution of RMSE differences highlights regions where FuXiWeather2 outperforms HRES (blue), regions where HRES performs better (red), and regions where the performance of both systems is comparable (white). For $R$, $T$, $U$, and $V$, FuXiWeather2 consistently exhibits lower RMSE values across nearly all map regions and all forecast lead times. For $Z$, the forecast advantages of FuXiWeather2 are primarily concentrated in low-latitude regions on the first day and gradually extends toward the mid- and high-latitude regions as the forecast lead time increases. Fig \ref{fig:sup_fcst_map_surface} presents the spatial comparisons for surface variables, including $MSLP$, $T_{2M}$, and $U_{10m}$. FuXiWeather2 demonstrates superior forecasting performance on a nearly global scale.

Figs. \ref{fig:sup_fcst_activity_upper} and \ref{fig:sup_fcst_activity_surf} presents RMSE-based activity, normalized relative to the forecast activity of the HRES. As a result, the HRES activity is scaled to 1 over the 10-day forecast period, with FuXi forecast activity values lower than 1 indicating reduced variability compared to HRES. Overall, the decrease in forecast activity for FuXiWeather2 occurs at a gradual and acceptable rate.
\subsection{Tracking tropical cyclones}
Tropical cyclones (TCs) are among the most destructive natural hazards, posing significant threats to coastal infrastructure and socioeconomic stability \cite{zhang2009tropical,yu2019impact}. We systematically evaluated tropical cyclone track forecasts from FuXiWeather2, HRES and GFS, with a focus on 6-hourly predictions within a 5-day forecast horizon. The analysis focused on 73 tropical cyclones occurring between July 2023 and June 2024, which are appeared in both the International Best Track Archive for Climate Stewardship (IBTrACS) datasets \cite{knapp2010international} and HRES forecasts. Track errors were calculated using the IBTrACS as the ground truth, which provides the most reliable estimates of tropical cyclone trajectories.

\begin{figure}[tbp]
  \centering
  \includegraphics[width=1.0\textwidth]{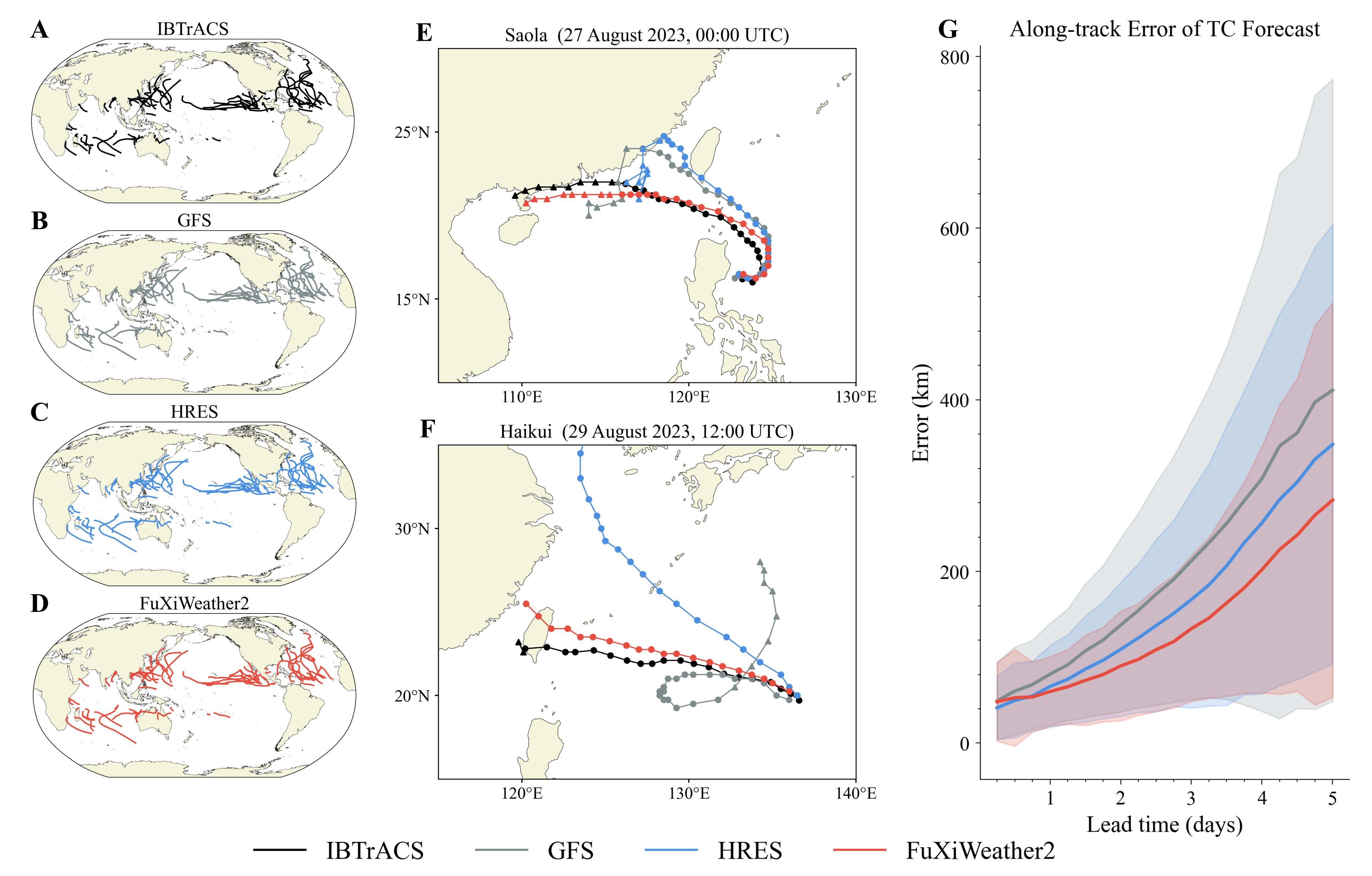}
  \caption{ 
  \textbf{Comparison of typhoon cyclone (TC) track forecasts.} (A) TC tracks in IBTrACS. (B–D) 24-hour TC track forecasts for GFS (B), HRES (C), and FuXiWeather2 (D). (E) TC track forecasts for Typhoon SAOLA, initialized at 00:00 UTC on August 27, 2023. (F) TC track forecasts for Typhoon HAIKUI, initialized at 12:00 UTC on August 29, 2023. (G) Along-track errors of TC track forecasts. Black, grey, blue, and red represent IBTrACS, GFS, HRES, and FuXiWeather2, respectively. The evaluation spans a one-year testing period, including forecasts initialized at 00:00 and 12:00 UTC with a 6-hour forecast step. In Panels E and F, The forecast lead times are truncated at the time of typhoon landfall. For Typhoon SAOLA (E), a 7-day forecast is presented, with lead times of 0–5 days denoted by dots and 5–7 days by triangles. For Typhoon HAIKUI (F), a 5-day forecast is displayed. In panel G, the shaded areas represent the standard deviation of the errors.}
  \label{fig:fcst_typhoon}
\end{figure}

Fig. \ref{fig:fcst_typhoon}A–D present the TC track of IBTrACS and 24-h TC track forecasts for GFS, HRES, and FuXiWeather2. As shown, the 24-hour forecast tracks of FuXiWeather2 exhibit high consistency with IBTrACS for these 73 tropical cyclones. Fig. 4E and 4F present the track forecast results for typhoon Saola and Haikui, respectively. The former is the second strongest typhoon to impact the South China Sea since 1950, while the latter set multiple historical rainfall records across the Guangdong-Hong Kong-Macao Greater Bay Area \cite{chan2025observational,wang2025microphysical}. FuXiWeather2 successfully predicted the landfall positions of both typhoons, whereas the forecast tracks from GFS and HRES exhibited significant deviations from the IBTrACS.

Based on the statistical results of track errors for all TCs, FuXiWeather2 outperforms both GFS and HRES (Fig. \ref{fig:fcst_typhoon}G). The along-track position errors of FuXiWeather2 at 3-day and 5-day lead times are 133.31 km and 283.18 km, respectively, representing reductions of 25.45\% and 22.89\% compared with the corresponding errors of 167.24 km and 348.00 km from HRES.
\subsection{Sensitivity Experiments of Simulated Observations}
\begin{figure}[ht]
\centering
\includegraphics[width=0.8\textwidth]{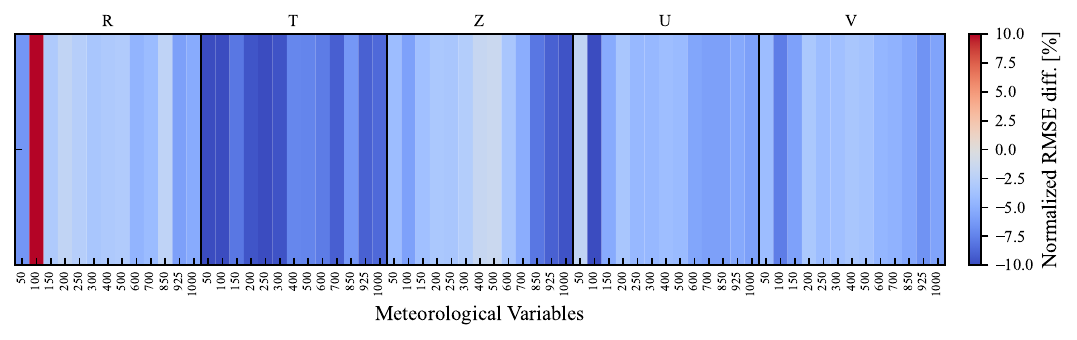}
\caption{
\textbf{Impact of simulated observations on analysis accuracy.} The heatmap displays the percentage change in root mean square error (RMSE) for the assimilation model trained on a hybrid dataset (comprising both real and simulated observations) relative to the baseline model (trained solely on real observations), with ERA5 used as the ground truth. The plot shows five upper-air variables across 13 pressure levels. The evaluation spans a one-year testing period at 00:00 and 12:00 UTC. Blue shading (negative values) indicates that the inclusion of simulated data results in a reduction of analysis errors.
}
\label{fig:sup_simulation_overall}
\end{figure}
\begin{figure}[ht]
\centering
\includegraphics[width=1.0\textwidth]{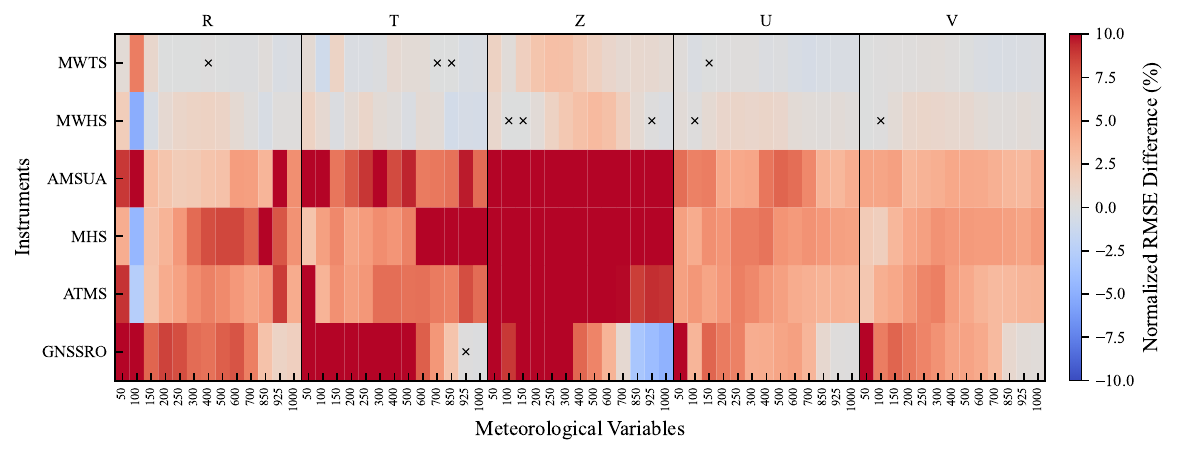}
\caption{
\textbf{Sensitivity analysis of satellite instruments for the assimilation model trained on real observations.} The heatmap displays the percentage change in root mean square error (RMSE) of the analysis fields after excluding a specific satellite instrument during the cyclic data assimilation and forecast phase, with ERA5 reanalysis used as the ground truth. The differences are calculated relative to a control experiment that assimilates all available satellite instruments. Red (blue) shading indicates a degradation (improvement) in performance following the data exclusion. Results cover five upper-air variables across 13 vertical pressure levels. The evaluation spans a one-year testing period at 00:00 and 12:00 UTC. The symbol $\times$ denotes cases where the difference is not statistically significant, based on the t-test at the 95\% confidence level.
}
\label{fig:sup_dde_wo_sim}
\end{figure}
\begin{figure}[h]
\centering
\includegraphics[width=1.0\textwidth]{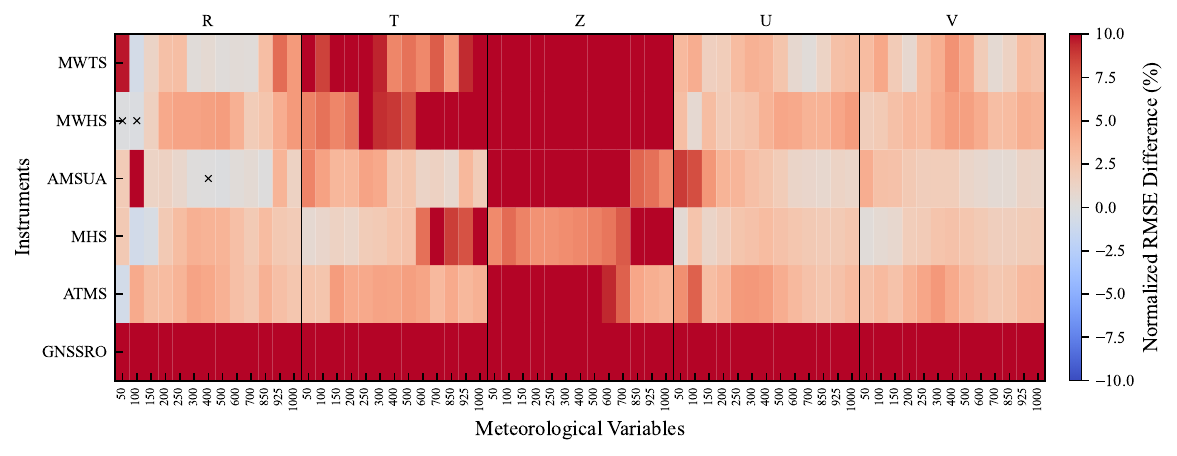}
\caption{
\textbf{Sensitivity analysis of satellite instruments for the assimilation model trained on hybrid observations.} The heatmap displays the percentage change in root mean square error (RMSE) of the analysis fields after excluding a specific satellite instrument during the cyclic data assimilation and forecast phase, with ERA5 reanalysis used as the ground truth. The differences are calculated relative to a control experiment that assimilates all available satellite instruments. Red (blue) shading indicates a degradation (improvement) in performance following the data exclusion. Results cover five upper-air variables across 13 vertical pressure levels. The evaluation spans a one-year testing period at 00:00 and 12:00 UTC. The symbol $\times$ denotes cases where the difference is not statistically significant, based on the t-test at the 95\% confidence level.
}
\label{fig:sup_dde_w_sim}
\end{figure}
We evaluated the overall performance shifts of the model after training with simulated data. Two pre-trained DA models were constructed for comparison: a baseline model trained exclusively on real satellite observations, and an enhanced model that integrates both real and simulated data. Specifically, the simulated data were primarily employed to fill observational gaps for instruments such as GNSS-RO, FY-3E MWHS-II, and FY-3E MWTS-III during the training period.

Fig. \ref{fig:sup_simulation_overall} illustrates the normalized RMSE difference between the analysis fields generated by the enhanced DA model and the baseline DA model, where both models assimilate the same satellite observations during the inference phase. The RMSE is calculated against the ERA5 reanalysis dataset. As indicated by the widespread blue shading across almost all meteorological variables ($R$, $T$, $Z$, $U$, $V$) and pressure levels, the model trained with simulated and real-world observation outperforms the baseline model. This demonstrates that filling the historical observational gaps with simulated data directly contributes to a significant improvement in analysis.

To investigate the underlying mechanism of this improvement, we conducted data denial experiments. This approach is widely used to assess the relative contributions of various observations in DA systems \cite{eyre2022part2,samrat2025observation}. During the cyclic assimilation and forecast process, specific satellite instruments were excluded to generate multiple sets of analysis fields. The contribution of each specific satellite instrument was quantified by calculating the percentage change in the RMSE of the resulting analysis fields relative to the analysis which assimilates all available data.

Fig \ref{fig:sup_dde_wo_sim} illustrates the normalized differences in the globally-averaged and latitude-weighted RMSE for the baseline model. As shown, the denial of any satellite observation generally leads to increased analysis errors compared to the control experiment. However, the impact of removing GNSS-RO, MWTS, and MWHS is significantly less pronounced than the findings in \cite{sun2025}, with some variables even exhibiting a decrease in RMSE. Compared to the baseline model, the enhanced model trained with both simulated and real-world data exhibits a much more pronounced performance degradation when GNSS-RO, MWHS, and MWTS data are missing (Fig. \ref{fig:sup_dde_w_sim}). For instance, the exclusion of GNSS-RO in the enhanced model results in a profound and widespread accuracy reduction across almost all meteorological variables ($R$, $T$, $Z$, $U$, $V$) and pressure levels. Similarly, the denial of MWHS and MWTS leads to darker red shading for variables such as temperature ($T$) and geopotential ($Z$), which indicates a more substantial increase in RMSE.

A plausible explanation for the anomalous performance of GNSS-RO, MWTS, and MWHS in the baseline model in the data denial experiments is that the DA model failed to sufficiently learn from observation modalities which are by frequently missing. By filling historical observational gaps with simulated data and establishing a uniform temporal distribution, the model can effectively enhance its feature extraction and fusion capabilities for these types of observations. However, since the simulated observations are idealized and do not adequately account for actual observational errors, the importance of GNSS-RO, MWTS, and MWHS appears disproportionately high in the enhanced model, which remains an issue to be addressed in the future. This experiment also serves as a reminder that the importance of various observations cannot be simply evaluated within a machine learning-based assimilation framework \cite{allen2025,laloyaux2025using,sun2025}; rather, it requires more equitable and comprehensive experimental settings.
\section{Discussion}
While highly efficient, most advanced neural weather prediction models \cite{pangu2023,graphcast,lang2024aifs,Chen2023} rely entirely on ERA5 as both their inputs and sole learning targets. Consequently, they inevitably inherit the systematic biases and latency of traditional DA pipelines. 
FuXiWeather2 breaks this "reanalysis emulation bottleneck". By establishing a fully differentiable assimilation-forecast cycle and incorporating high-fidelity real-world observations as supplementary supervision, our framework transitions neural weather prediction from dependent emulation to a fully autonomous, operational forecasting paradigm.

FuXiWeather2's analysis fields outperform both the gold-standard ERA5 reanalysis and the ECMWF HRES system in the lower troposphere and at the surface, where highly complex and rapidly changing atmospheric processes often occur. 
Due to the limited representation capabilities of parameterization schemes for land-surface exchange processes, traditional numerical methods inherently introduce significant systematic model errors during the assimilation of surface observations \cite{sandu2020addressing, de2022coupled, ingleby2024improved}.
FuXiWeather2 overcomes this limitation by utilizing neural networks to directly learn the assimilation of discrete surface observations from historical data. Driven by these superior analysis fields, FuXiWeather2's forecasts surpass those of HRES in 91\% of evaluated metrics, underscores its practical utility.

Despite these advances, FuXiWeather2 has several limitations. First, its coarser resolution (25 km, 13 levels vs. HRES’s 9 km, 137 levels) restricts its ability to capture fine-scale atmospheric features. Second, the analysis accuracy at higher atmospheric levels still lags behind HRES. Incorporating additional upper-air observations (e.g., geostationary satellite data and aircraft reports) is necessary to close this gap. Third, similar with other regression-based models \cite{pangu2023,graphcast,Chen2023,allen2025,sun2025,gupta2026healda}, FuXiWeather2’s forecasts exhibit lower activity compared to NWP systems (see materials and methods), which inherently leads to an underestimation of extreme event intensities \cite{bonavita2024some,subich2025fixing}. Refining the loss functions or transitioning from deterministic regression models to generative probabilistic models is warranted \cite{Price2025,subich2025fixing}. 

Methodologically, the current system relies on empirical parameters to weight the uncertainties between observational and reanalysis labels. As future iterations scale to incorporate a broader diversity of observational modalities, such empirical heuristics will become inherently unreliable. Developing data-driven approaches to autonomously model these uncertainties represents a highly promising solution \cite{kendall2018multi,kendall2017uncertainties}. A similar heuristic limitation exists regarding observational non-stationarity: we currently mitigate the observational distribution inconsistency by simulating historical data via physical forward models. While data denial experiments confirm this strategy's overall efficacy, future work should explore advanced generative models to simulate higher-fidelity observations and incorporate transfer learning techniques to robustly bridge the simulation-to-reality gap.

Looking beyond these methodological refinements, future efforts aim to transition FuXiWeather2 from an atmosphere-centric focus to a fully coupled multi-sphere framework. By integrating these interacting spheres into a unified neural architecture \cite{wang2024coupled}, we will explore the cross-sphere assimilation of multi-source observations (e.g., sea surface temperature and soil moisture). Ultimately, this paradigm will pave the way for a consistent and holistic simulation of the entire Earth system \cite{de2022coupled,boucher2025learning}.
\section*{Acknowledgments}
We thank the European Centre for Medium-Range Weather Forecasts (ECMWF) for providing the ERA5 reanalysis data as well as the HRES analysis and forecast data.
We also thank the National Oceanic and Atmospheric Administration (NOAA) for providing the GFS analysis and forecast data, the GDAS observational data, and the ICOADS data.
We also express our gratitude to the China Meteorological Administration (CMA) and Tianjin Yunyao Aerospace Technology Co., Ltd. for providing the observational data.
\bibliographystyle{unsrtnat}
\bibliography{references}  

\renewcommand{\thefigure}{S\arabic{figure}}
\setcounter{figure}{0}
\renewcommand{\theHfigure}{Appendix.\thefigure}

\begin{figure}[p]
  \centering
  \includegraphics[height=0.8\textheight]{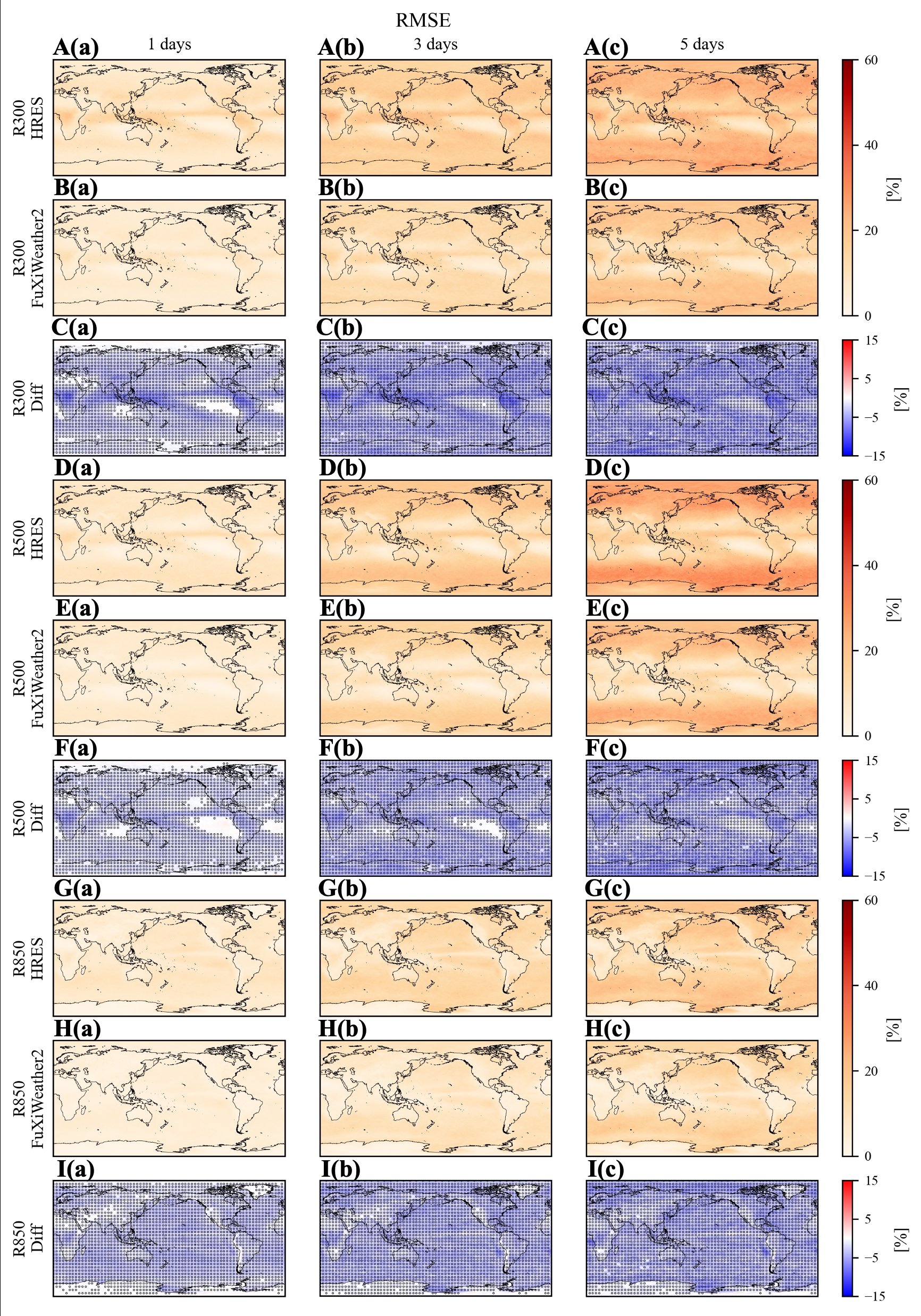}
  \caption{ \textbf{Spatial distribution of time-averaged root mean square error (RMSE) for relative humidity ($R$).} 
  (A-C) RMSE of HRES (A) and FuXiWeather2 (B) forecasts, along with their differences (C) at 300 hPa. 
  (D-F) RMSE of HRES (D) and FuXiWeather2 (E) forecasts, along with their differences (F) at 500 hPa. 
  (G-I) RMSE of HRES (G) and FuXiWeather2 (H) forecasts, along with their differences (I) at 850 hPa.
  a, b and c (columns 1, 2, and 3) represent 1-day, 3-day, and 5-day lead times, respectively.
  the evaluation spans a one-year testing period, including forecasts initialized at 00:00 and 12:00 UTC.
  In RMSE maps, darker red indicates higher errors; in difference maps, blue (red) denotes regions where FuXiWeather2 (HRES) performs better. 
  In panels C, F and I, stippling highlights regions where FuXiWeather2 significantly outperforms HRES (t-test, 95\% confidence level)
  }
  \label{fig:sup_fcst_map_r}
\end{figure}

\begin{figure}[p]
  \centering
  \includegraphics[height=0.8\textheight]{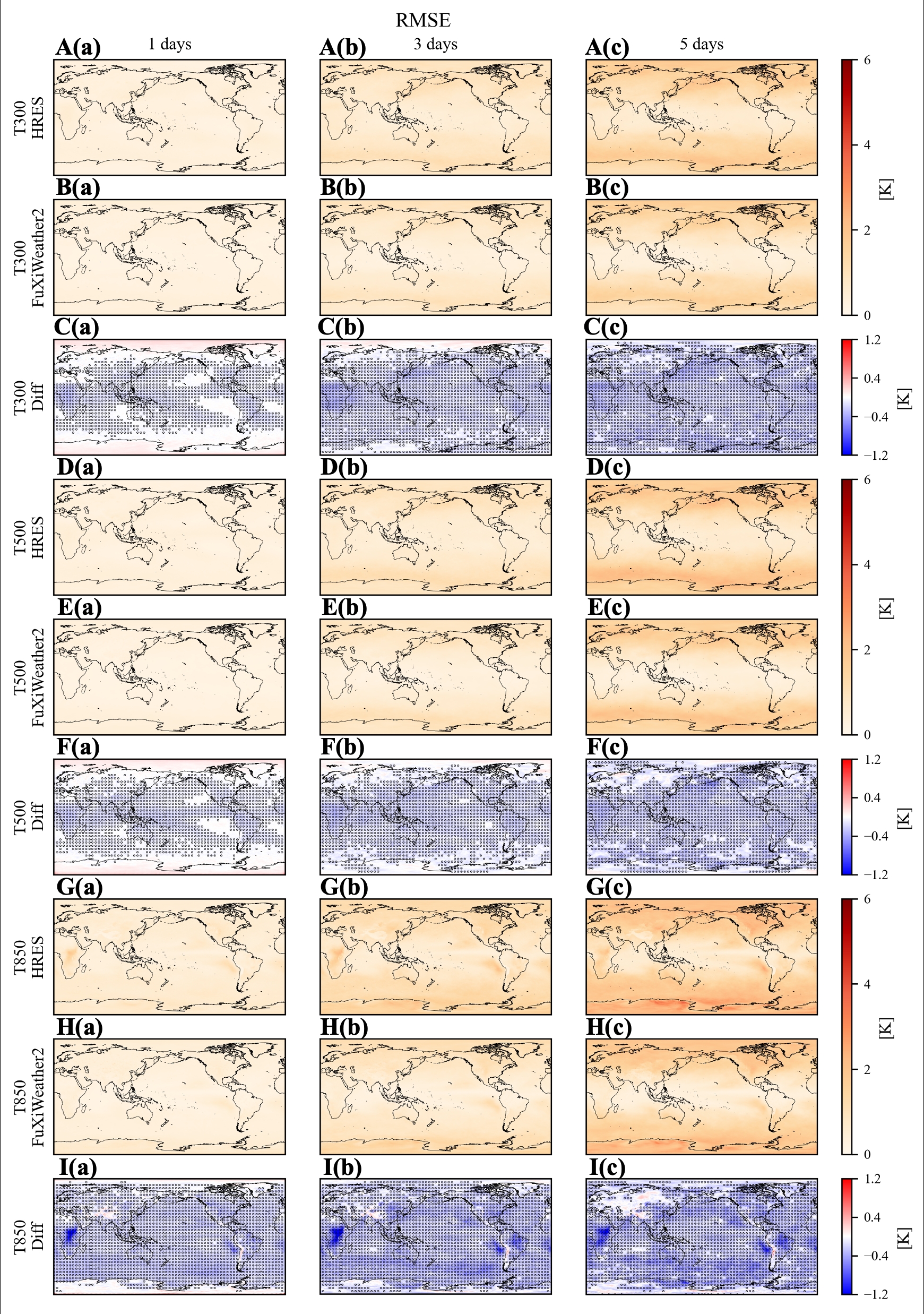}
  \caption{ \textbf{Spatial distribution of time-averaged root mean square error (RMSE) for temperature ($T$).} 
  (A-C) RMSE of HRES (A) and FuXiWeather2 (B) forecasts, along with their differences (C) at 300 hPa. 
  (D-F) RMSE of HRES (D) and FuXiWeather2 (E) forecasts, along with their differences (F) at 500 hPa. 
  (G-I) RMSE of HRES (G) and FuXiWeather2 (H) forecasts, along with their differences (I) at 850 hPa.
  a, b and c (columns 1, 2, and 3) represent 1-day, 3-day, and 5-day lead times, respectively.
  the evaluation spans a one-year testing period, including forecasts initialized at 00:00 and 12:00 UTC.
  In RMSE maps, darker red indicates higher errors; in difference maps, blue (red) denotes regions where FuXiWeather2 (HRES) performs better. 
  In panels C, F and I, stippling highlights regions where FuXiWeather2 significantly outperforms HRES (t-test, 95\% confidence level)
  }
  \label{fig:sup_fcst_map_t}
\end{figure}

\begin{figure}[p]
  \centering
  \includegraphics[height=0.8\textheight]{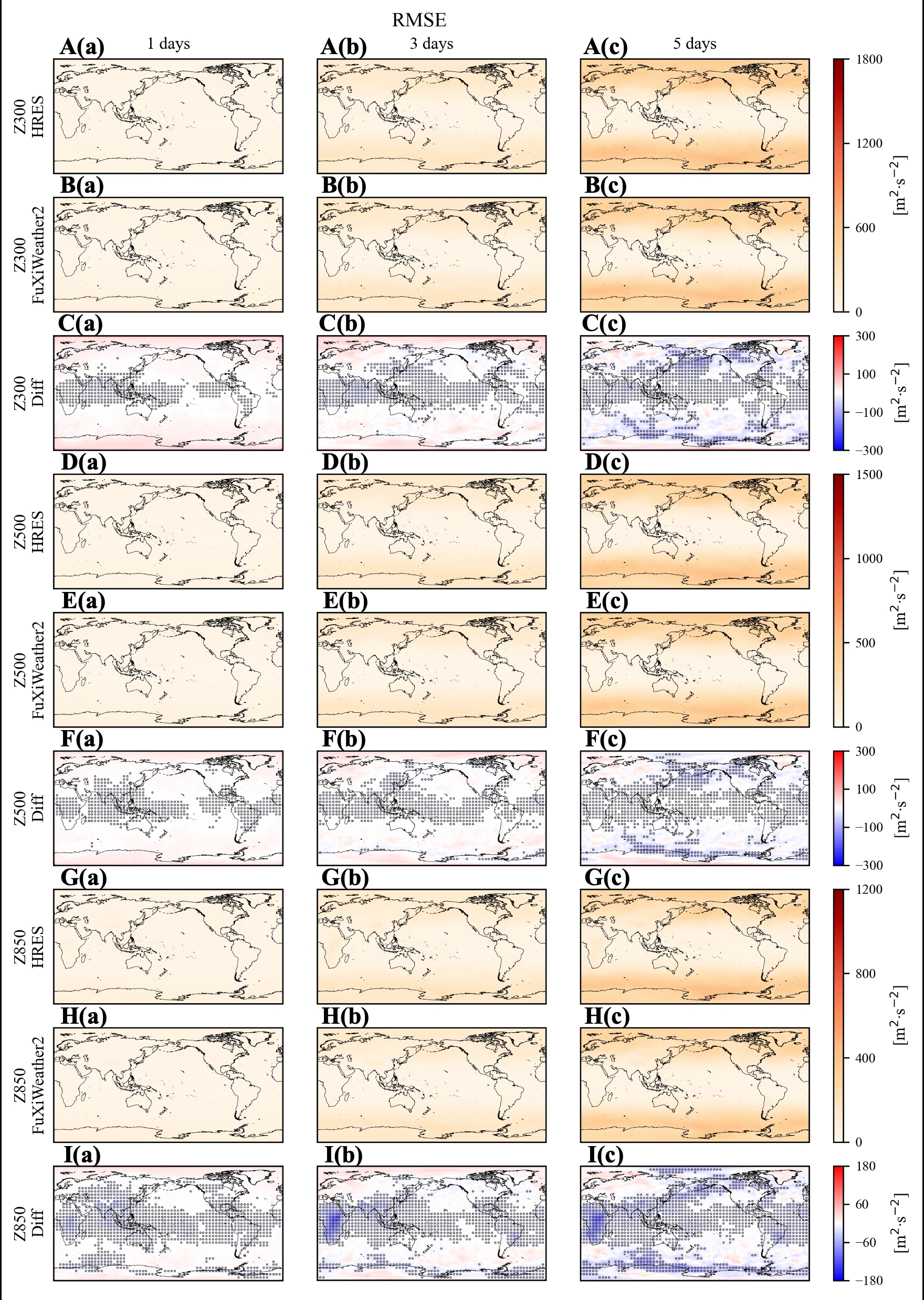}
  \caption{ \textbf{Spatial distribution of time-averaged root mean square error (RMSE) for geopotential ($Z$).} 
  (A-C) RMSE of HRES (A) and FuXiWeather2 (B) forecasts, along with their differences (C) at 300 hPa. 
  (D-F) RMSE of HRES (D) and FuXiWeather2 (E) forecasts, along with their differences (F) at 500 hPa. 
  (G-I) RMSE of HRES (G) and FuXiWeather2 (H) forecasts, along with their differences (I) at 850 hPa.
  a, b and c (columns 1, 2, and 3) represent 1-day, 3-day, and 5-day lead times, respectively.
  the evaluation spans a one-year testing period, including forecasts initialized at 00:00 and 12:00 UTC.
  In RMSE maps, darker red indicates higher errors; in difference maps, blue (red) denotes regions where FuXiWeather2 (HRES) performs better. 
  In panels C, F and I, stippling highlights regions where FuXiWeather2 significantly outperforms HRES (t-test, 95\% confidence level)
  }
  \label{fig:sup_fcst_map_z}
\end{figure}

\begin{figure}[p]
  \centering
  \includegraphics[height=0.8\textheight]{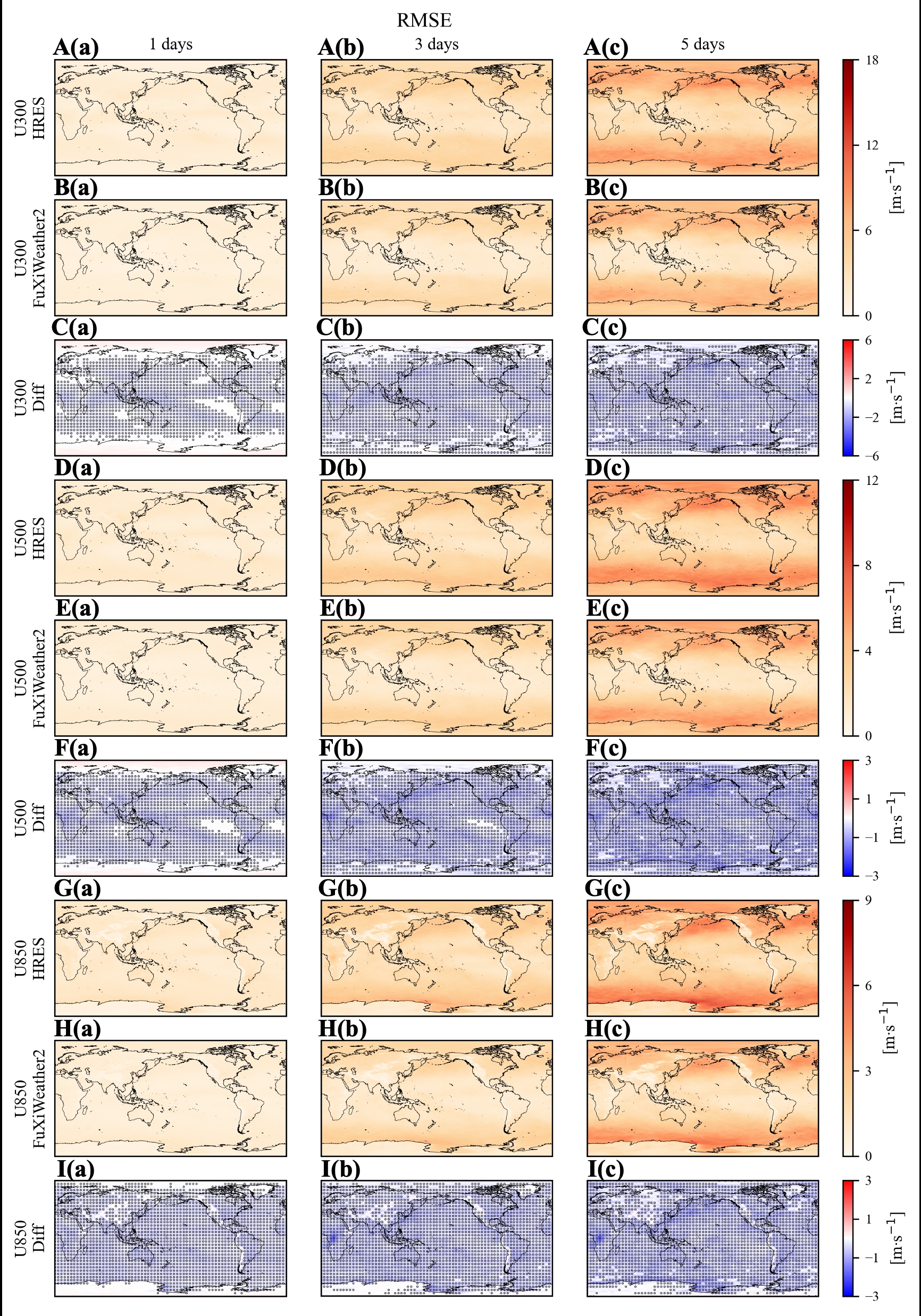}
  \caption{ \textbf{Spatial distribution of time-averaged root mean square error (RMSE) for U-wind component ($U$).} 
  (A-C) RMSE of HRES (A) and FuXiWeather2 (B) forecasts, along with their differences (C) at 300 hPa. 
  (D-F) RMSE of HRES (D) and FuXiWeather2 (E) forecasts, along with their differences (F) at 500 hPa. 
  (G-I) RMSE of HRES (G) and FuXiWeather2 (H) forecasts, along with their differences (I) at 850 hPa.
  a, b and c (columns 1, 2, and 3) represent 1-day, 3-day, and 5-day lead times, respectively.
  the evaluation spans a one-year testing period, including forecasts initialized at 00:00 and 12:00 UTC.
  In RMSE maps, darker red indicates higher errors; in difference maps, blue (red) denotes regions where FuXiWeather2 (HRES) performs better. 
  In panels C, F and I, stippling highlights regions where FuXiWeather2 significantly outperforms HRES (t-test, 95\% confidence level)
  }
  \label{fig:sup_fcst_map_u}
\end{figure}

\begin{figure}[p]
  \centering
  \includegraphics[height=0.8\textheight]{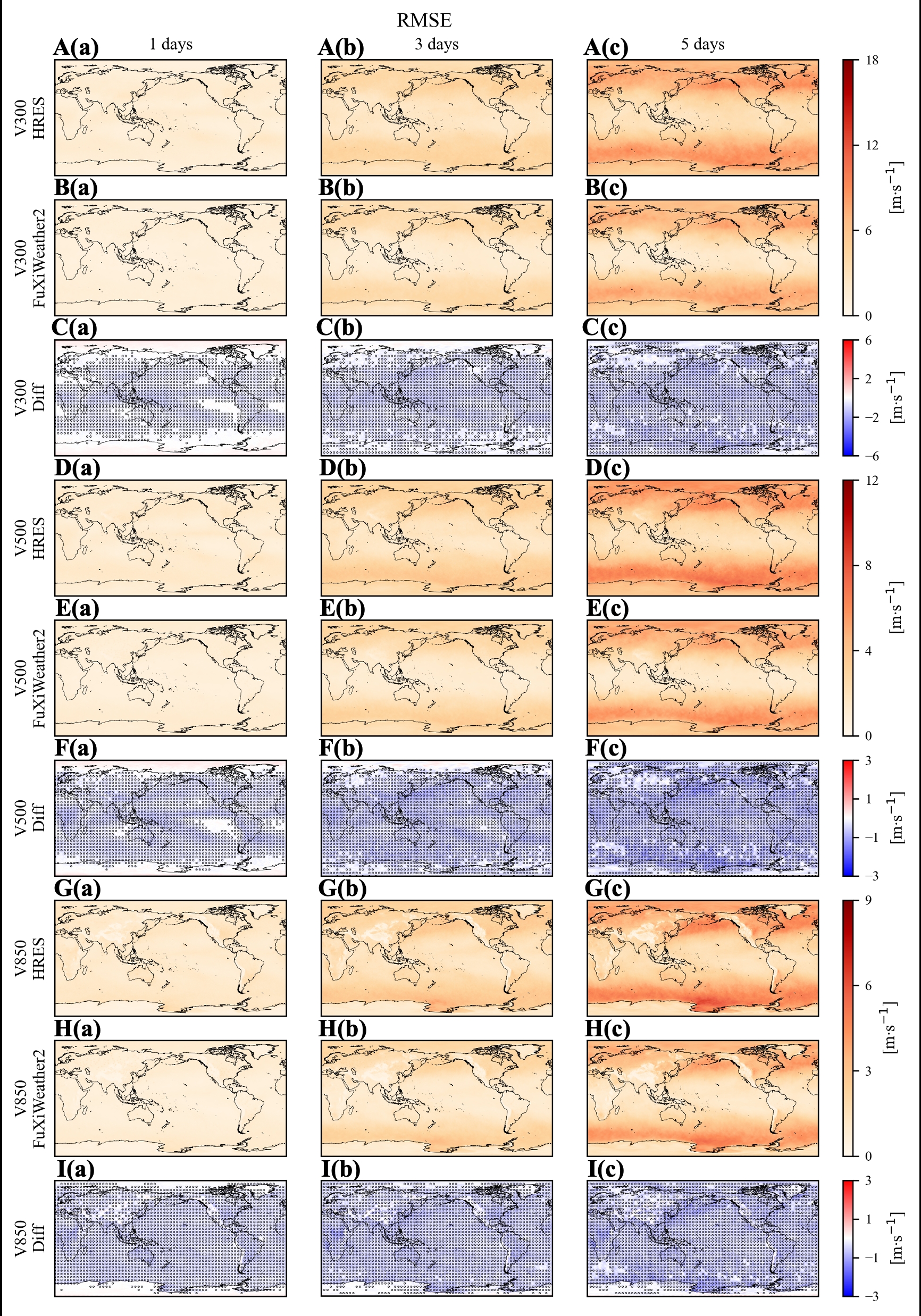}
  \caption{ \textbf{Spatial distribution of time-averaged root mean square error (RMSE) for V-wind component ($U$).} 
  (A-C) RMSE of HRES (A) and FuXiWeather2 (B) forecasts, along with their differences (C) at 300 hPa. 
  (D-F) RMSE of HRES (D) and FuXiWeather2 (E) forecasts, along with their differences (F) at 500 hPa. 
  (G-I) RMSE of HRES (G) and FuXiWeather2 (H) forecasts, along with their differences (I) at 850 hPa.
  a, b and c (columns 1, 2, and 3) represent 1-day, 3-day, and 5-day lead times, respectively.
  the evaluation spans a one-year testing period, including forecasts initialized at 00:00 and 12:00 UTC.
  In RMSE maps, darker red indicates higher errors; in difference maps, blue (red) denotes regions where FuXiWeather2 (HRES) performs better. 
  In panels C, F and I, stippling highlights regions where FuXiWeather2 significantly outperforms HRES (t-test, 95\% confidence level)
  }
  \label{fig:sup_fcst_map_v}
\end{figure}

\begin{figure}[p]
  \centering
  \includegraphics[height=0.8\textheight]{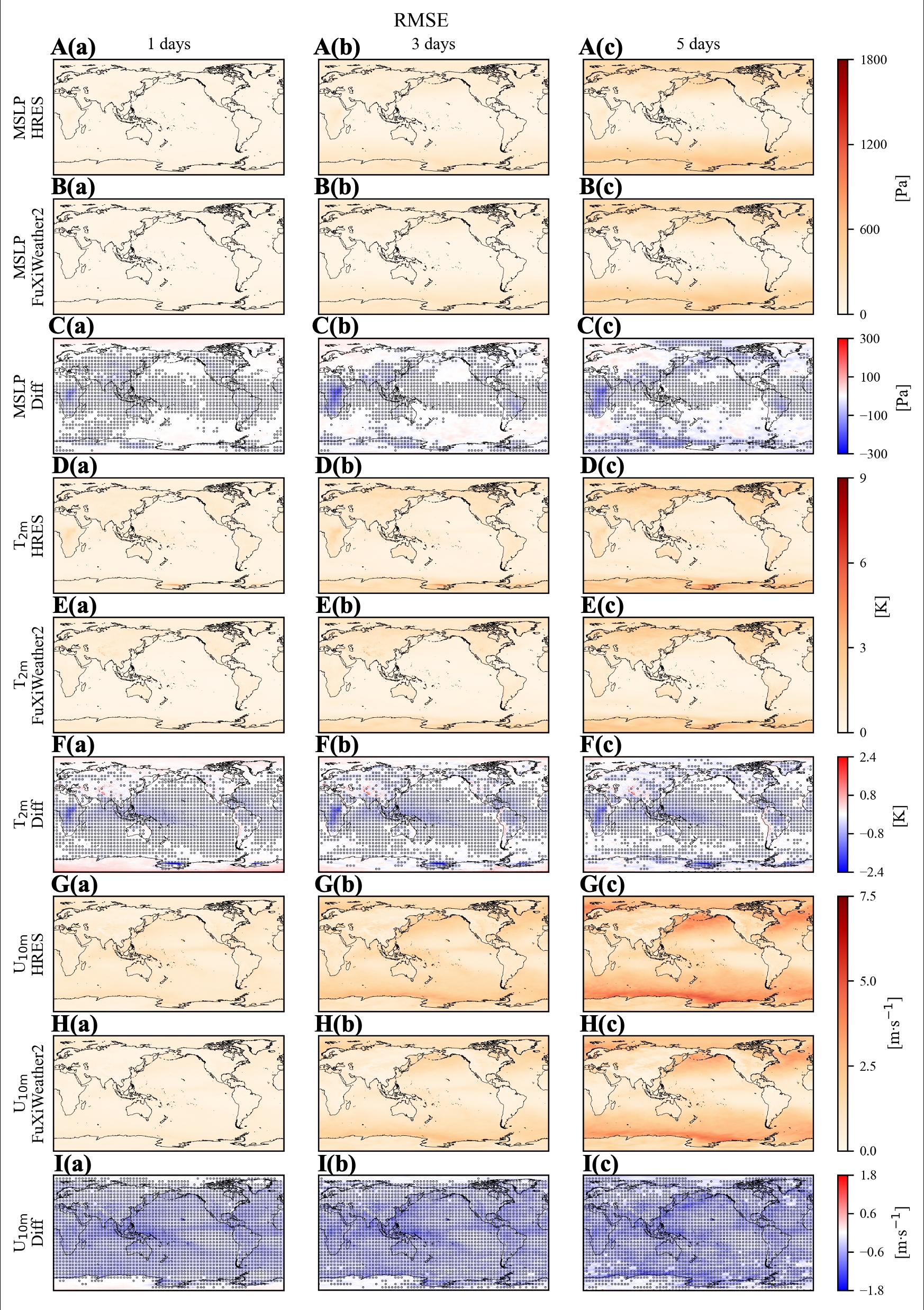}
  \caption{ \textbf{Spatial distribution of time-averaged root mean square error (RMSE) for surface variables.} 
  (A-C) RMSE of HRES (A) and FuXiWeather2 (B) forecasts, along with their differences (C) for mean sea-level pressure ($MSLP$). 
  (D-F) RMSE of HRES (D) and FuXiWeather2 (E) forecasts, along with their differences (F) for 2-meter temperature ($T_{2m}$). 
  (G-I) RMSE of HRES (G) and FuXiWeather2 (H) forecasts, along with their differences (I) for 10-meter U-wind component ($U_{10m}$).
  a, b and c (columns 1, 2, and 3) represent 1-day, 3-day, and 5-day lead times, respectively.
  the evaluation spans a one-year testing period, including forecasts initialized at 00:00 and 12:00 UTC.
  In RMSE maps, darker red indicates higher errors; in difference maps, blue (red) denotes regions where FuXiWeather2 (HRES) performs better. 
  In panels C, F and I, stippling highlights regions where FuXiWeather2 significantly outperforms HRES (t-test, 95\% confidence level)
  }
  \label{fig:sup_fcst_map_surface}
\end{figure}

\end{document}